\documentclass[10pt, logo, copyright]{nvidiatechreport}

\usepackage{mdframed}
\usepackage[utf8]{inputenc} 
\usepackage[T1]{fontenc}    

\usepackage{amsfonts}       
\usepackage{nicefrac}       
\usepackage{microtype}      
\usepackage[dvipsnames]{xcolor}         
\usepackage{multirow}
\usepackage{multicol}
\usepackage{graphicx}
\usepackage[numbers]{natbib}
\usepackage{tabto}
\usepackage{xspace}
\usepackage{amsmath}
\usepackage{adjustbox}
\usepackage{enumitem}
\usepackage{wrapfig}
\usepackage{dblfloatfix}

\usepackage{footmisc}
\usepackage{listings}

\usepackage{float}

\usepackage{hyperref}
\usepackage{array}
\usepackage{ragged2e}
\usepackage{subcaption}
\usepackage{caption}

\usepackage{natbib}

\def\eqref#1{equation~\ref{#1}}

\def\1{\bm{1}}

\DeclareMathAlphabet{\mathsfit}{\encodingdefault}{\sfdefault}{m}{sl}
\SetMathAlphabet{\mathsfit}{bold}{\encodingdefault}{\sfdefault}{bx}{n}

\usepackage{amsmath}
\usepackage{amssymb}
\usepackage{mathtools}
\usepackage{amsthm}
\usepackage{enumitem}
\setlist[itemize]{itemsep=0.5em}
\usepackage{multirow}
\usepackage{booktabs}

\usepackage[capitalize,noabbrev]{cleveref}

\usepackage{algorithm}
\usepackage{algorithmicx}
\usepackage{algpseudocode}
\usepackage{xspace}
\usepackage{csquotes}
\usepackage{listings}
\usepackage{xcolor}
\definecolor{NvidiaGreen}{rgb}{0,0.392,0}

\usepackage{multirow}
\usepackage{multicol}
\usepackage{enumitem}

\usepackage[most]{tcolorbox}
\newcommand\minisection[1]{\vspace{1.3mm}\noindent \textbf{#1}}

\newcommand{\finding}[2]{
    \begin{tcolorbox}[
        colback=white!90!gray,
        colframe=teal!60!black,
        arc=5pt,
        boxsep=5pt,
        left=10pt,
        right=10pt,
        top=2pt,
        bottom=2pt,
        boxrule=0.8pt,
        drop shadow=gray!0!white,
        enhanced jigsaw
    ]
        \minisection{Key Finding #1:} #2
    \end{tcolorbox}
}

\newcommand{\insight}[2]{
    \begin{tcolorbox}[
        colback=white!90!gray,
        colframe=teal!60!black,
        arc=5pt,
        boxsep=5pt,
        left=10pt,
        right=10pt,
        top=2pt,
        bottom=2pt,
        boxrule=0.8pt,
        drop shadow=gray!0!white,
        enhanced jigsaw
    ]
        \minisection{Key Insight #1:} #2
    \end{tcolorbox}
}

\definecolor{lightgray}{gray}{0.95} 
\definecolor{darkblue}{rgb}{0,0,0.6} 
\definecolor{nvgreen}{cmyk}{50, 0, 100, 0}

\title{DLER: Doing Length pEnalty Right - Incentivizing More Intelligence per Token via Reinforcement Learning}

\author{Shih-Yang Liu\textonesuperior, Xin Dong\textsuperscript{*}, Ximing Lu, Shizhe Diao, Mingjie Liu, Min-Hung Chen, Hongxu Yin, Yu-Chiang Frank Wang, Kwang-Ting Cheng\textonesuperior, Yejin Choi, Jan Kautz, Pavlo Molchanov  \\~\\ \textbf{NVIDIA} \\ \ }
\correspondingauthor{X}
\begin{abstract}
Reasoning language models such as OpenAI-o1, DeepSeek-R1, and Qwen achieve strong performance via extended chains of thought but often generate unnecessarily long outputs. Maximizing intelligence per token—accuracy relative to response length—remains an open problem. We revisit reinforcement learning (RL) with the simplest length penalty—truncation—and show that accuracy degradation arises not from the lack of sophisticated penalties but from inadequate RL optimization. We identify three key challenges: (i) large bias in advantage estimation, (ii) entropy collapse, (iii) sparse reward signal. We address them with \textbf{D}oing \textbf{L}ength p\textbf{E}nalty \textbf{R}ight (\textbf{DLER}), a training recipe combining batch-wise reward normalization, higher clipping, dynamic sampling, and simple truncation length penalty. DLER achieves state-of-the-art accuracy–efficiency trade-offs, cutting output length by over 70\% while surpassing all previous baseline accuracy. It also improves test-time scaling: compared to DeepSeek-R1-7B, DLER-7B generates multiple concise responses in parallel with 28\% higher accuracy and lower latency. We further introduce Difficulty-Aware DLER, which adaptively tightens truncation on easier questions for additional efficiency gains. We also propose an update-selective merging method that preserves baseline accuracy while retaining the concise reasoning ability of the DLER model which is useful for scenarios where RL training data is scarce.

\end{abstract}

\begin{document}
\maketitle

\noindent\textbf{Models on Hugging Face: \href{https://huggingface.co/collections/nvidia/reasoning-tokens-efficiency-68a8ea0ffe21f3fc46e1da0f}{Reasoning Token Efficiency}}

\section{Introduction}

\begin{figure}[h]
\begin{center}
\centering
\vspace{-10pt}
\begin{subfigure}{0.48\textwidth}
  \centering
  \includegraphics[width=\linewidth]{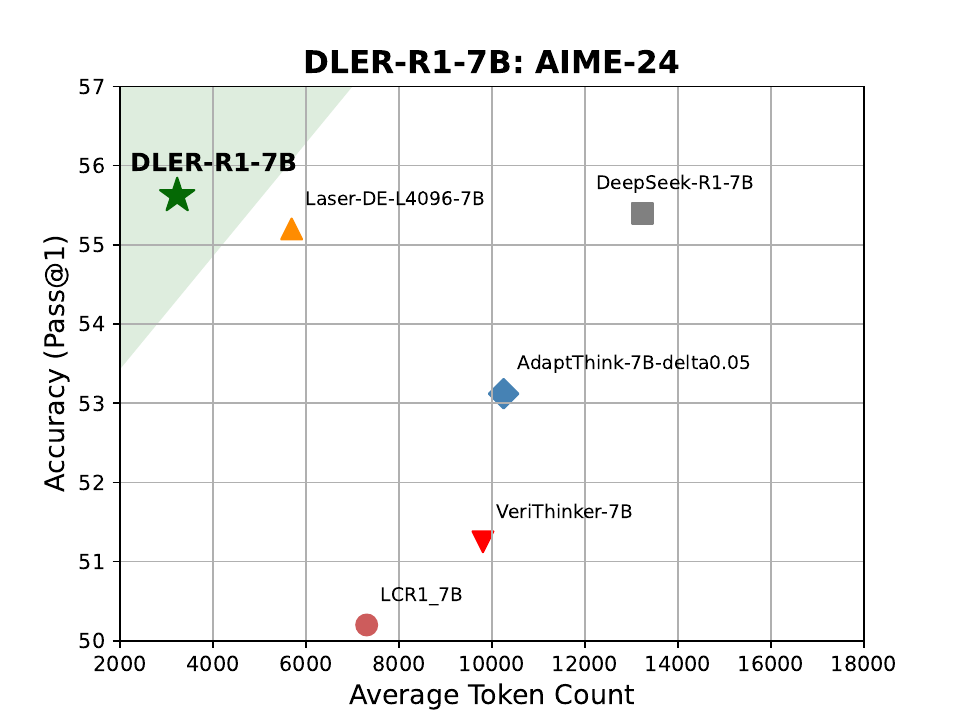}
  \caption{DLER Training on DeepSeek-R1-7B}
  \label{fig:teaser}
\end{subfigure}
\hfill
\begin{subfigure}{0.48\textwidth}
  \centering
  \includegraphics[width=\linewidth]{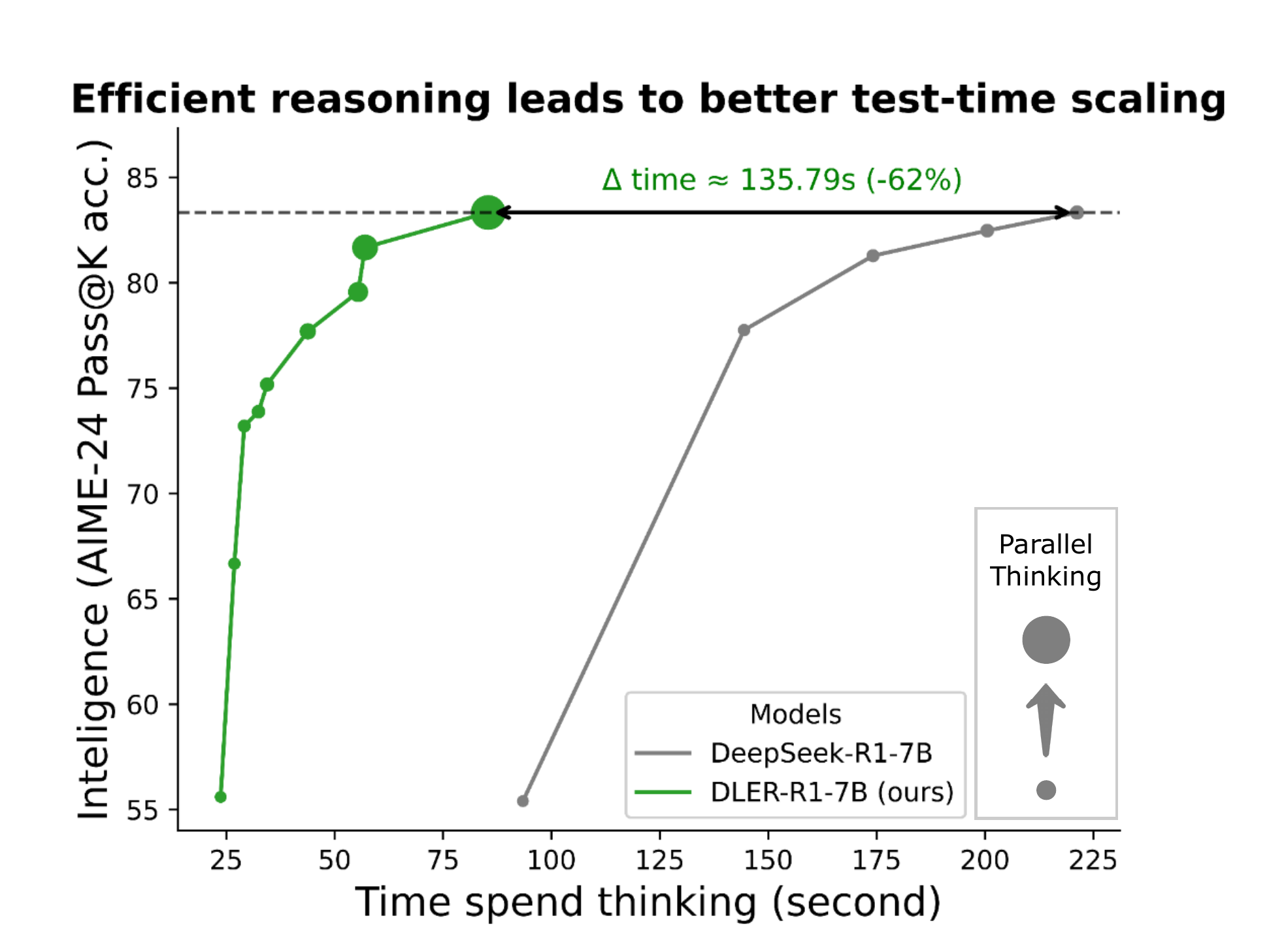}
  \caption{DLER-R1-7B enable better test-time scaling}
  \label{fig:parallel_latency_teaser}
\end{subfigure}
\end{center}
\vspace{-10pt}
\caption{(a) DLER achieves state-of-the-art accuracy/length trade-offs, shortening CoT by up to 70\% without losing accuracy. (b) On AIME-24, DLER-R1 models enable better test-time scaling. Results for 1.5B models are shown in Fig.~\ref{fig:teaser_appendix} and Fig.~\ref{fig:parallel_latency_teaser_appendix} in the appendix.} 
\end{figure}

Reasoning models such as OpenAI-o1~\citep{jaech2024openai}, DeepSeek-R1~\citep{guo2025Deepseek}, and Qwen~\citep{qwen3} achieve strong performance through long chains of thought (CoT)~\citep{yang2025speculativethinking}, but this comes at the cost of heavy token usage, higher latency, and redundant outputs for questions solvable with shorter responses. Therefore, how to maximize intelligence per token remains an open research question. Recent work has addressed the inefficiency of extended reasoning by developing methods to reduce output length. These approaches fall into three categories: prompt engineering~\citep{ma2025nothink}, supervised fine-tuning~\citep{Lu2025RetroSearchEU, chen2025verithinker,liu2024canskip,xia2025tokenskip, ma2025cotvalve}, and reinforcement learning (RL)~\citep{fang2025thinkless,liu2025laser,luo2025o1pruner,hou2025thinkprune,aggarwal2025l1}. Among these, RL-based methods have emerged as the most principled approach for achieving optimal accuracy-efficiency trade-offs. These methods typically incorporate length penalties into the reward function to incentivize reasoning within predefined token budgets. However, despite demonstrating substantial reductions in reasoning length, existing approaches often suffer from accuracy degradation that varies significantly across tasks of different complexity levels—a limitation we hypothesize stems from suboptimal optimization techniques.

In this work, \textbf{we revisit reinforcement learning (RL) for reasoning efficiency by re-examining the simplest length penalty—truncation}, which assigns zero reward to responses exceeding a fixed limit. Prior RL methods using truncation often fail to recover accuracy; we find this stems not from the length penalty itself but from sub-optimal RL optimization techniques. Three issues drive this degradation: (i) biased advantage estimation under Group Relative Policy Optimization (GRPO)~\citep{shao2024DeepSeekmath} due to substantial reward noise, especially in early state of training, as many responses are abruptly cut off and assigned zero reward, (ii) persistent entropy collapse that hampers exploration of diverse reasoning paths, and (iii) sparse reward signals arising from a large portion of prompts in each batch where all rollouts are truncated and thus assigned zero reward. We address these by adopting batch-wise normalization~\citep{hu2025reinforce++}, higher clipping thresholds~\citep{yu2025dapo}, to promote exploration via low-probability,
high-entropy tokens, and curriculumized filtering to gradually introduce harder prompts~\cite{yu2025dapo}.

By combining all essential elements, our final training recipe, \textit{\textbf{D}oing \textbf{L}ength p\textbf{E}nalty \textbf{R}ight (\textbf{DLER})}, achieves state-of-the-art accuracy-to-token efficiency. 
As illustrated in Fig.~\ref{fig:teaser}, DLER fully recovers the accuracy drop while reducing the average response length by over 70\%. This underscores key insights: 

\insight{1}{It is not the sophisticated design of the length penalty that determines performance, but rather the choice of RL optimization algorithm. Even the simplest length truncation can achieve state-of-the-art accuracy-to-token efficiency when combined with our DLER recipe. See Sec.~\ref{sec:main_results} for more details.}

\insight{2}{We additionally apply DLER recipe to a variety of length penalties and find that they no longer push the frontier of accuracy-efficiency frontier, but instead serve as tools for fine-grained adjustment of the trade-offs. See Sec.~\ref{sec:diff_length_peanlties} for more details.}

We also benchmark test-time scaling by generating multiple responses in parallel. Fig.~\ref{fig:parallel_latency_teaser} shows that our DLER-R1-7B delivers significant improvements over the original DeepSeek-R1-7B, achieving a 27\% accuracy gain (AIME-24) within the same wall-clock “thinking time.”  This marks an important shift in perspective: whereas recent efforts~\citep{guo2025Deepseek,liu2025prorl,yu2025dapo} to enhance reasoning ability have largely pursued accuracy through increasingly long reasoning traces, our findings demonstrate that:

\insight{3}{Improving reasoning efficiency not only lowers the cost of single response but also enables superior test-time parallel scaling. See Sec.~\ref{sec:parallel_thinking} for more details.}

We further propose a difficulty-aware variant of DLER (\textbf{DA-DLER}), where the truncation target length is dynamically adjusted according to an estimate of the model’s ability to solve the question. For questions that the model can already reliably answer within the target length, the target length is further shortened to encourage even shorter reasoning, while more challenging questions are allowed more tokens. DA-DLER can achieve an additional 15\% and 11\% reduction in response length on DeepSeek-R1-1.5B and 7B, respectively, further advancing the efficiency frontier.

We also provide a solution for practical scenarios where accessing the original RL training dataset is not feasible. Often, applying small-scale academic RL training datasets to proprietary models leads to accuracy degradation, and length penalties exacerbate this issue~\citep{liu2025laser} while remaining effective at reducing output length. To completely mitigate accuracy degradation without access to the original dataset, we adopt an update-selective weight merging strategy that combines the original baseline model with the DLER-trained model. This method recovers all lost accuracy while still reducing output tokens by 47\%. 

\insight{4}{Weight merging allows for a better trade-off between accuracy and length reduction especially when accessing the original high-quality proprietary datasets is not an option. See Sec.~\ref{sec:weight_merging} for more details.}

In summary, our contributions are as follows:
\begin{itemize}
\item  We revisit the simplest length penalty—truncation—and identify the fundamental factors behind the poor accuracy reported in previous studies - the inadequate RL optimization techniques. 

\item  Building on these insights, we carefully integrate effective RL optimization techniques to address these issues. This allows us to achieve state-of-the-art accuracy-to-length ratios using only the simplest truncation penalty, significantly outperforming prior methods that rely on more complex length penalties for enhancing the reasoning efficiency of DeepSeek-R1-1.5B and 7B. Our findings reveal a key takeaway: improvements in the accuracy–efficiency trade-off are influenced less by the design of the penalty function and more by the choice of optimization algorithm. 

\item  We introduce a difficulty-aware extension of DLER that adaptively adjusts truncation length based on model capability, shortening truncation length for easy questions and relaxing them for harder ones, achieving an additional 15\% and 11\% reduction in response length on DeepSeek-R1-1.5B and 7B.

\item RL training with small-scale datasets cuts response length by 55\% but causes an accuracy drop on some tasks due to insufficient data. We address this with a update-selective weight merging strategy that recovers the accuracy while still being able to reduce the average output length by 47\%, offering a training-free path to accurate and efficient reasoning models without high-quality proprietary data.

\end{itemize}

\section{Preliminary}

Reinforcement learning is widely applied to enhance the reasoning ability of modern LMs~\cite{comanici2025gemini, achiam2023gpt}, with GRPO~\citep{shao2024DeepSeekmath} becoming popular for its efficiency in removing the critic model and using group-relative advantage estimation. This approach maintains token-level advantage estimation accuracy while significantly reducing the overhead. Specifically, for each question-answer pair \((q, a)\), the behavior policy \(\pi_{\theta_{\mathrm{old}}}\) samples a group of \(G\) responses \(\{o_i\}_{i=1}^G\). The advantage for the \(i\)-th response is then computed by normalizing the group-level rewards \(\{R_i\}_{i=1}^G\) as:
\begin{equation}
A_{i,t} = \frac{R_i - \mathrm{mean}(\{R_i\}_{i=1}^G)}{\mathrm{std}(\{R_i\}_{i=1}^G)}
\label{eq:grpo_advantage}
\end{equation}

and the optimization objective is formulated as: 
\begin{equation}
\mathcal{J}_{\mathrm{GRPO}}(\theta) = 
\mathbb{E}_{(q,a) \sim D,\; \{o_i\}_{i=1}^G \sim \pi_{\theta_{\mathrm{old}}}(\cdot|q)} 
\left[
\frac{1}{G} \sum_{i=1}^G \frac{1}{|o_i|} \sum_{t=1}^{|o_i|}
\min \left(
s_{i,t}(\theta)\, A_{i,t},\;
\mathrm{clip}(s_{i,t}(\theta), 1 - \epsilon, 1 + \epsilon)\, A_{i,t}
\right)
\right]
\label{eq:grpo}
\end{equation}
where $s_t(\theta) = \frac{\pi_{\theta}(o_t \,|\, q, o_{<t})}{\pi_{\theta_{\mathrm{old}}}(o_t \,|\, q, o_{<t})}$ and $\epsilon$ is the clipping threshold. We omit the KL Loss for simplicity. Recent studies~\citep{liu2025laser,fang2025thinkless,hou2025thinkprune,aggarwal2025l1}, aiming to enhance reasoning efficiency, commonly adopt GRPO as the optimization algorithm, 
coupled with custom-designed length penalty rewards. Under this setup, the new reward \( R'_i \) 
is generally formulated as: $R'_i = R_i + L_i$ where \(R_i\) denotes the correctness reward, computed using rule-based heuristics, and 
\(L_i\) represents the length penalty. Depending on the specific length penalty, \(L_i\) may either impose a larger penalty on longer outputs or, in the case of the simple truncation penalty, assign a reward of zero 
to any response that exceeds a predefined length limit.

\section{Re-examining the Simplest Length Penalty - Truncation} 

Prior studies that aim to enhance training efficiency tend to treat the underlying policy optimization algorithm as a fixed, reliable component, often attributing improvements in accuracy-to-length ratio primarily to the design of length penalties. 
However, this overlooks the possibility that the optimization algorithm itself may introduce performance bottlenecks. 
In this work, we re-examine the structure of the policy optimization objective by adopting the simplest possible length penalty—truncation—which is simple enough to alleviate reward hacking and enables a focused analysis of how the optimization algorithm alone affects accuracy degradation.

\subsection{More Aggressive Truncation Leads to Higher Reward Variance} 

\begin{figure}[h]
\begin{center}
\includegraphics[width=\textwidth]{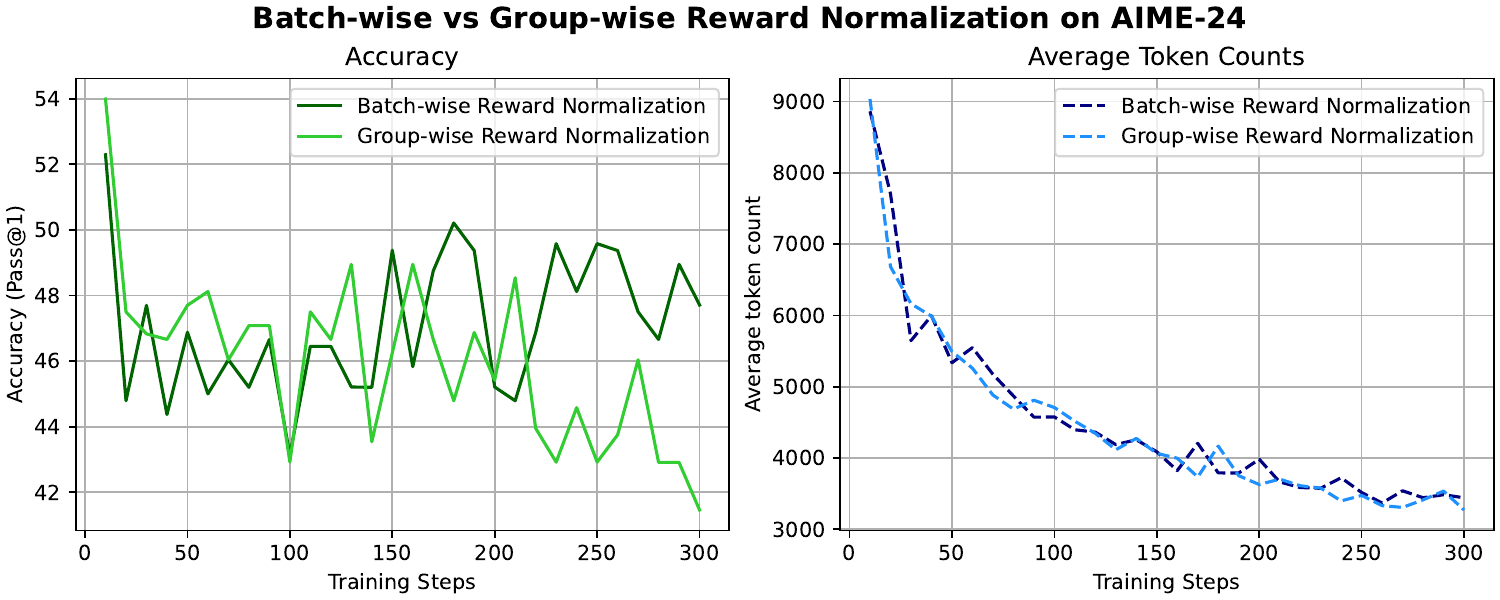}
\end{center}
\caption{Accuracy and average response length of DeepSeek-R1-7B on the AIME-24 test set, evaluated every 10 training steps across two RL training runs: group-wise reward normalization (GRPO) and batch-wise normalization. GRPO shows declining accuracy while batch-wise reward normalization remains stable despite reduced token counts.}
\label{fig:reinforce-grpo-baseline}
\end{figure}

First, we find that truncation greatly increases reward variance.
To assess this, we calculate the advantage variance of DeepSeek-R1-7B at step 0 using 16 rollouts per question, and average the results over a batch of 512 questions drawn from the DeepScaleR-Preview-Dataset~\cite{deepscaler2025}. Testing truncation lengths of \{4000, 8000, 12000, 16000\}, we observe a clear pattern: more aggressive truncation lengths lead to higher per-prompt variance on average, with corresponding values of \{0.4, 0.32, 0.3, 0.29\}. These results suggest that truncation introduces greater training instability by increasing advantage variance, which in turn leads to more biased advantage estimates according to Equ.~\ref{eq:grpo_advantage}—a topic we elaborate on further in the Appendix.\ref{appendix:larger_bias}. 

To address the increased bias introduced by truncation, we propose replacing GRPO's prompt-wise advantage normalization with global batch-wise advantage normalization, which is also used in~\citep{hu2025reinforce++}. The idea is that by normalizing the advantages across the entire batch, we can mitigate the impact of outliers and ensure a more stable estimation of advantage variance. The advantage calculation for the $i$-th response after adopting batch-wise reward normalization becomes:
\begin{equation}
A^{\mathrm{norm}}_{i, t} = 
\frac{A_{i, t} - \mathrm{mean}_{\mathrm{batch}}(A_{i, t})}
     {\mathrm{std}_{\mathrm{batch}}(A_{i, t})}
\label{eq:normalized-advantage}
\end{equation}
where $A_{i, t} = R'_{i} - \mathrm{mean}(\{R'_i\}_{i=1}^G)$ and we can see that the normalization is now done on batch level rather than local group level. 

We compare GRPO and batch-wise reward normalization by training DeepSeek-R1-7B on the DeepScaleR-Preview-Dataset and evaluate every 10 steps on AIME-24; see Sec.~\ref{sec:setup} for full experimental details. As illustrated in Fig.~\ref{fig:reinforce-grpo-baseline}, both GRPO and batch-wise reward normalization progressively shorten the average output length over training. 
However, GRPO shows a continuous drop in accuracy, whereas batch-wise reward normalization begins to recover accuracy after approximately 100 steps and can improve the accuracy of GRPO by around 3\% using approximately the same number of tokens.
This supports our earlier observation that GRPO's instability under length truncation leads to degraded performance, while switching to batch-wise reward normalization helps stabilize training and restore accuracy.

\finding{1}{Length truncation increases reward variance, leading to biased advantage estimates and degraded performance when using GRPO. Switching to batch-wise reward variance estimation mitigates this issue and improves performance.}

\subsection{Entropy Collapse Limits Exploration of Reasoning Paths} 
\label{sec:entropy_collapse}



\begin{figure}[ht]
  \centering
  \vspace{-10pt}
  \begin{subfigure}[b]{0.38\textwidth}
    \centering
    \includegraphics[width=\textwidth]{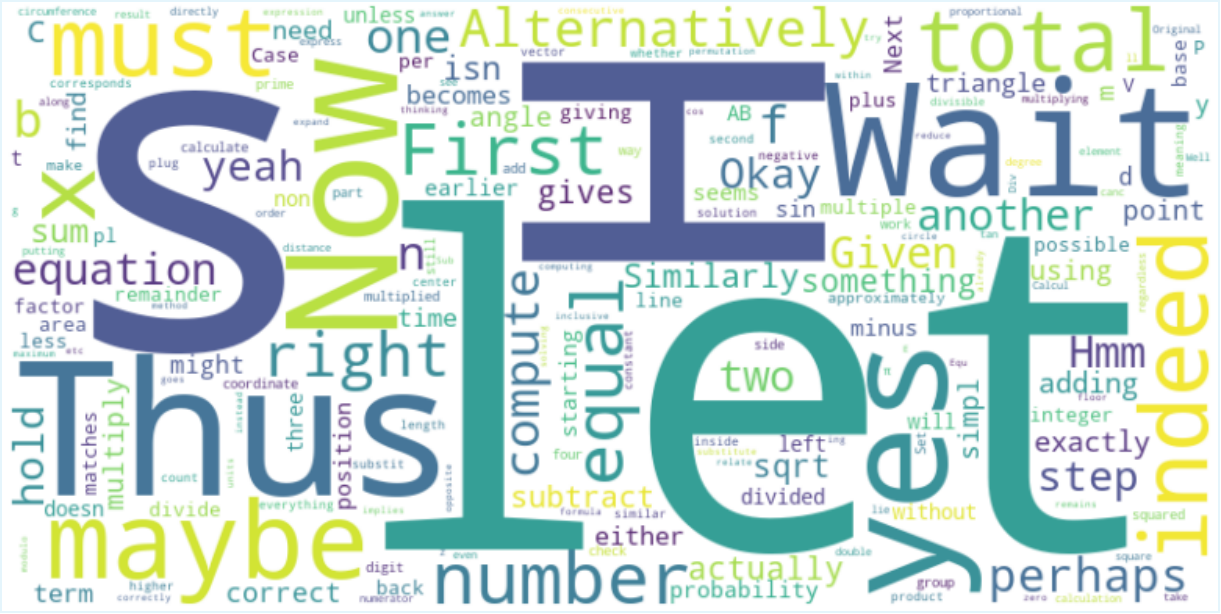}
    \vspace{0.2pt}
    \caption{}
    \label{fig:high-clip-word-cloud}
  \end{subfigure}
  \begin{subfigure}[b]{0.6\textwidth}
    \centering
    \includegraphics[width=\textwidth]{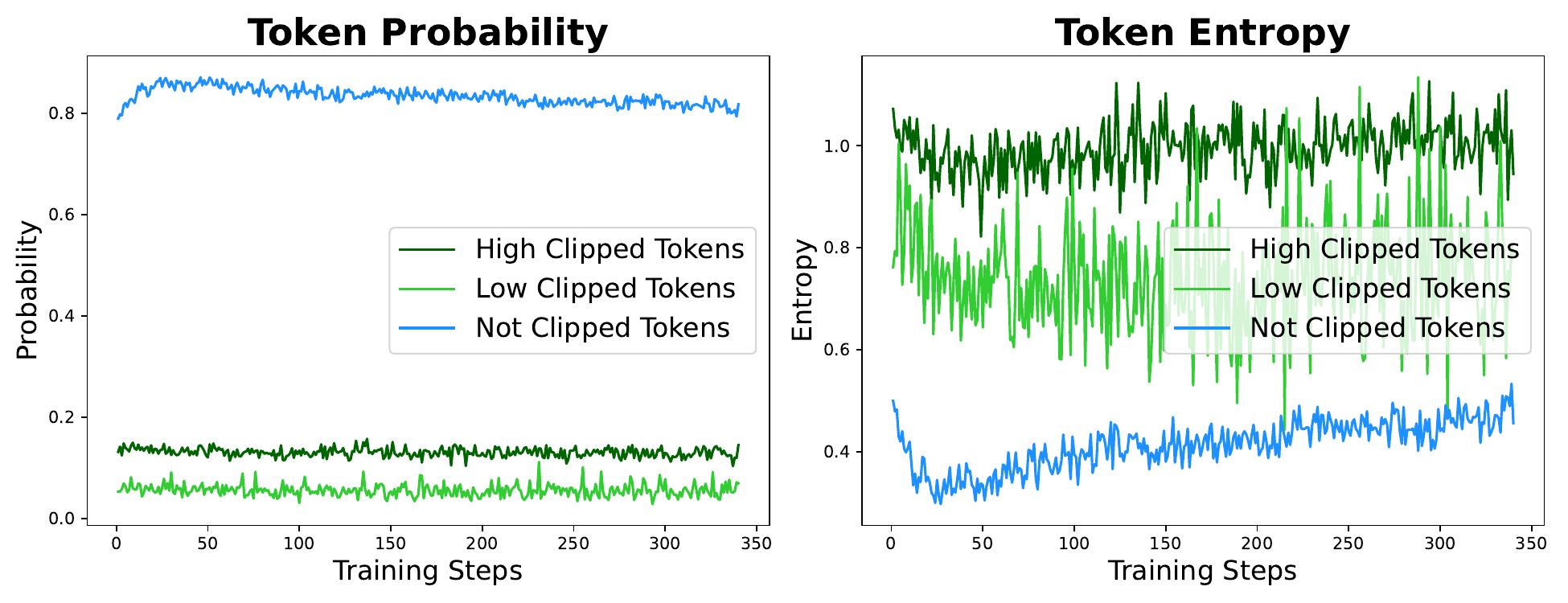}
    \caption{}
    \label{fig:clip-prob-entropy}
  \end{subfigure}
  \vspace{-5pt}
  \caption{\textbf{(a)} Word clouds of the most frequent tokens clipped by the high-threshold ($1+\epsilon$) before applying higher clipping threshold,showing that many are transitional words important for reasoning, and clipping them limits exploration during RL training.. \textbf{(b)} Average probability and entropy of tokens not clipped, clipped by the higher threshold, and clipped by the lower threshold during RL training. Clipped tokens have much lower probabilities than unclipped ones, and those clipped by the higher threshold consistently show higher entropy, supporting Fig.~\ref{fig:high-clip-word-cloud} that these are often high-entropy transitional tokens triggering reasoning steps.}
\label{fig:entropy_clopase_analysis}
\end{figure}

Although switching from group-wise reward normalization to the batch-wise reward normalization improves accuracy, it still falls short of fully restoring the reasoning model's performance. We observe that the current policy optimization still suffers from entropy collapse—a phenomenon identified in recent studies~\cite{yu2025dapo, liu2025prorl} as detrimental to learning. When entropy collapses, the model’s output distribution becomes overly concentrated, causing the policy to prematurely focus on a narrow set of responses. This limits exploration, introduces bias in policy updates, and ultimately stalls training progress.

We suspect that the entropy collapse issue may be caused by the clipping on the importance sampling ratio in Eq.~\ref{eq:grpo}. If one token is clipped, its gradient is effectively zeroed out, preventing it from contributing to the policy update. 
So, we are interested in understanding which tokens are most frequently clipped during training, and whether these tokens play a crucial role in exploration of either correct reasoning paths or controlling lengths. 

We visualize word clouds of the clipped tokens, as shown in Fig.~\ref{fig:high-clip-word-cloud}. 
Even they are only about 1\% of the total tokens, these tokens are largely composed of words such as “Wait,” “Hmm,” “Alternatively” (signaling contrast or shifts), and “thus” or “also” (indicating progression or causality), which often function as transitional cues in the model’s reasoning paths~\cite{wang2025beyond}. Moreover, these tokens play an important role in determining response length, since a higher frequency of such tokens generally corresponds to longer reasoning sequences.

More interestingly, we find there is a big overlap between low-probability tokens, high-entropy tokens, and clipped tokens.
In Fig.~\ref{fig:clip-prob-entropy}, we plot the average token probability and entropy for three groups: 1) tokens that are not clipped, 2) tokens clipped by the high-threshold ($1+\epsilon$), and 3) those clipped by the low-threshold ($1-\epsilon$). We find that clipped tokens—especially those clipped by the upper threshold—tend to have lower probabilities and higher entropy simultaneously. 
This finding also connects studies on low-probability tokens~\cite{yang2025not} and high-entropy tokens~\cite{wang2025beyond} and provides an unified perspective to understand previous different findings on token importance for optimization. 

By decoupling the lower and upper thresholds—previously set to the same value—and assigning a larger value to the upper threshold ($\epsilon_{high}$), the gradient update on those high entropy exploratory tokens are retained, allowing their gradients to propagate and fostering more diverse reasoning behaviors during training.

We conduct experiments on DeepSeek-R1-7B comparing batch-wise reward normalization with and without the higher clipping strategy, under a truncation target length of 4000. We visualize the corresponding average response length and average entropy in the training batch throughout training in Fig.~\ref{fig:training_dynamic_a} and Fig.~\ref{fig:training_dynamic_b}, where we observe that enabling higher clipping threshold clearly alleviates entropy collapse: the entropy of the model’s output distribution not only avoids vanishing, but even increases after an initial drop—contrasting with the behavior observed in the original DAPO~\cite{yu2025dapo} and ProRL~\cite{liu2025prorl} paper without the truncation length penalty.


\finding{2}{The clipped tokens are often low-probability, high-entropy tokens that play a crucial role in exploration of reasoning paths and length control. Adopting a higher clipping threshold helps retain these tokens in gradient updates, thereby mitigating entropy collapse.}


\subsection{Length Penalty Over-sparsify Training Signal}
\begin{figure}[h]
\begin{center}
\includegraphics[width=\textwidth]{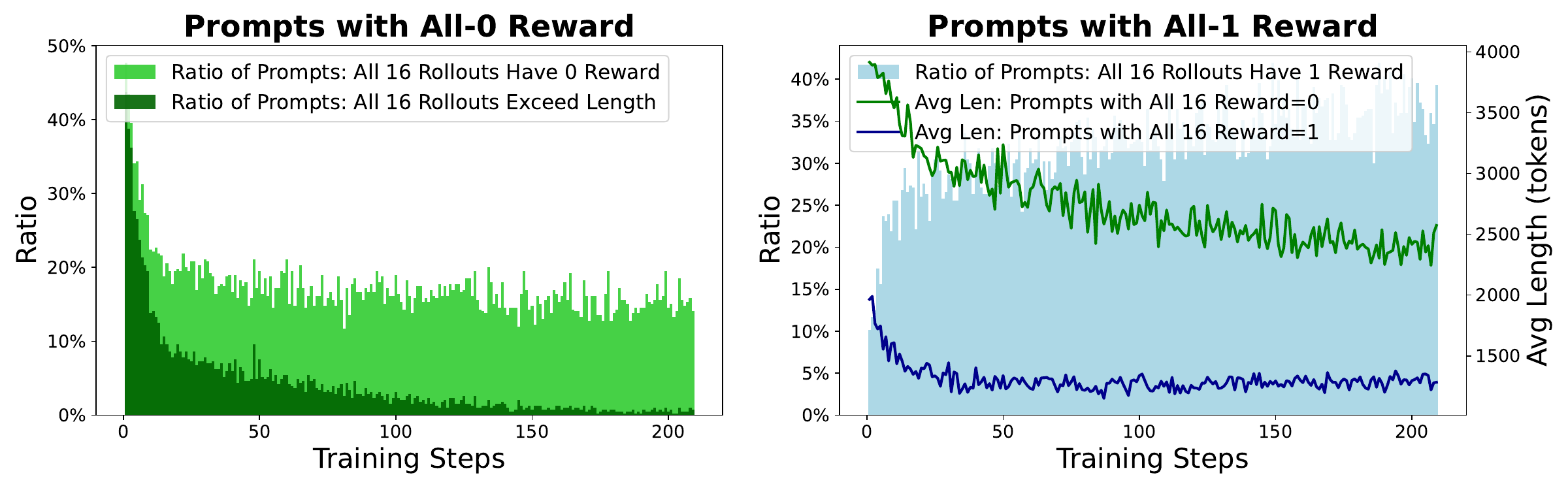}
\end{center}
\caption{\textbf{Left}: Ratio of training prompts with all 16 rollouts receiving zero reward, including those caused by exceeding the truncation length. Around half of the prompts fall into this category early in training, weakening the signal and biasing the model toward easier prompts that model already know how to solve within the target length. \textbf{Right}: Ratio of training prompts with all 16 rollouts receiving reward score of one steadily increases, while average response length declines and remains markedly shorter than that for prompts whose rollouts all receive a reward of zero.}
\label{fig:ds_zero_reward_ratio}
\end{figure}

Another phenomenon we identify with the application of length penalty is the prevalence of zero-reward signals across training rollouts. Specifically, a substantial fraction of the prompts receive zero reward for all 16 rollouts, primarily because all 16 responses exceed the target length. 

As illustrated in Fig.~\ref{fig:ds_zero_reward_ratio}, at the start of training nearly half of all prompts are affected by this issue. 
Such sparse and noisy feedback biases the model toward shorter, easier prompts it can already solve, thereby reducing exploration and constraining effective learning on improving accuracy. 

On the other hand, in the middle and later phases of training, the proportion of prompts with all rollouts receiving positive reward increases their domination, consisting of near 40\% of the batch as shown in Fig.~\ref{fig:ds_zero_reward_ratio}.
These prompts are typically easier, allowing the model to consistently generate quite short responses to answer them correctly.
However, when such prompts dominate, the model overfits to over-shorter responses and fails to fully utilize the target length budget.
This explains why a suboptimal RL training run with this issue (Fig.~\ref{fig:training_dynamic_b}) plateaus at a very short response length of 2k tokens despite the maximum length being set to 4k tokens, as the model has prematurely overfitted to these easy prompts.

In summary, both challenging prompts that receive all zero rewards and easy prompts that receive all positive rewards can skew the training distribution, leading to suboptimal learning outcomes.
To mitigate this issue, we discard prompts whose rollouts all yield zero reward or all yield positive reward and resample until the target batch size is reached \cite{yu2025dapo}. This dynamic sampling strategy implicitly induces a curriculum, as it progressively incorporates harder examples that initially demand longer reasoning chains to solve. 
As evident from the training dynamics shown in Fig.~\ref{fig:training_dynamic_b} and \ref{fig:training_dynamic_c}, the model autonomously learns to rapidly reduce token usage in the early training stages, then gradually increases length to fully exploit the target token budget. This behavior is driven purely by the simple truncation length penalty. Without dynamic sampling, the model tends to plateau at a shorter length, as shown in Fig.~\ref{fig:training_dynamic_b}, indicating that it has prematurely overfitted into a suboptimal local minimum.

\finding{3}{Dynamic sampling filters out prompts either too easy or too hard and responses either too long or too short, either of which can over-dominate the training batch and skew the reward shape. This leads to a more balanced training signal and enables the model to better utilize the target length budget.}

\subsection{Combining All Ingredients: Do Length pEnalty Right}

\begin{figure}[h]
\begin{center}
\includegraphics[width=\textwidth]{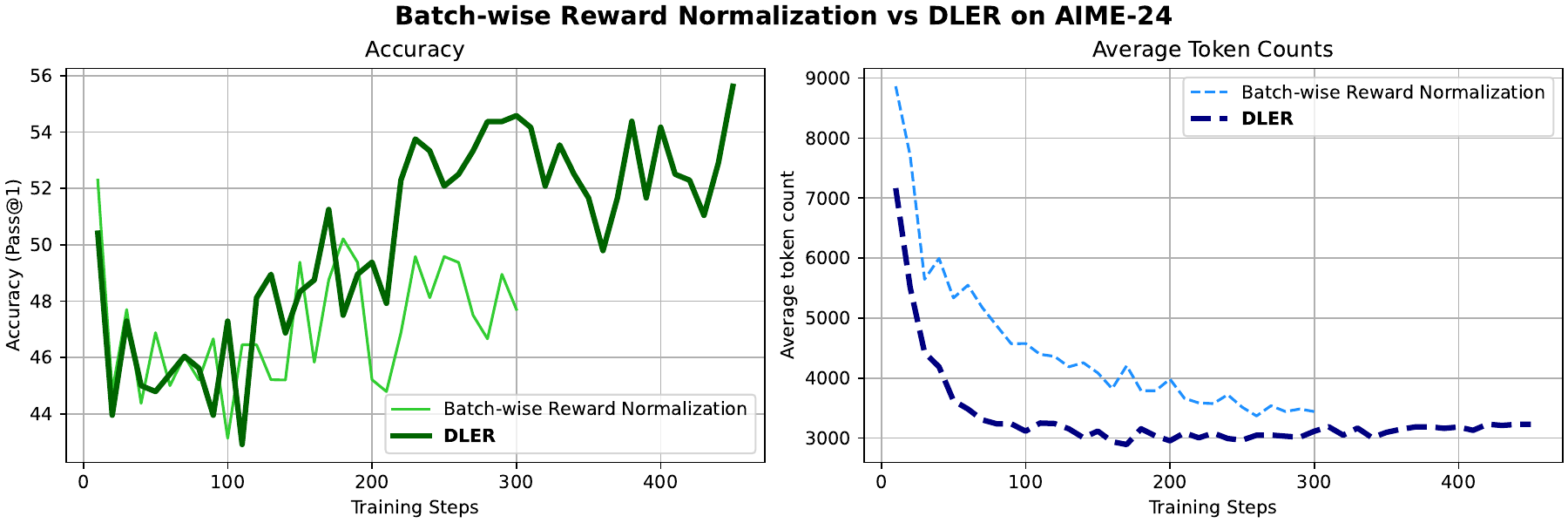}
\end{center}
\caption{Accuracy and average response length of DeepSeek-R1-7B on the AIME-24 test set, evaluated every 10 training steps across two RL training runs: one with batch-wise reward normalization and the other with our DLER recipe. DLER enhances the accuracy–token efficiency by reducing token usage while fully recovering the accuracy loss of applying plain batch-wise reward normalization.}
\label{fig:our_aime}
\end{figure}

\begin{figure}[h]
  \centering
  \begin{subfigure}[b]{0.45\textwidth}
    \centering
    \includegraphics[width=\textwidth]{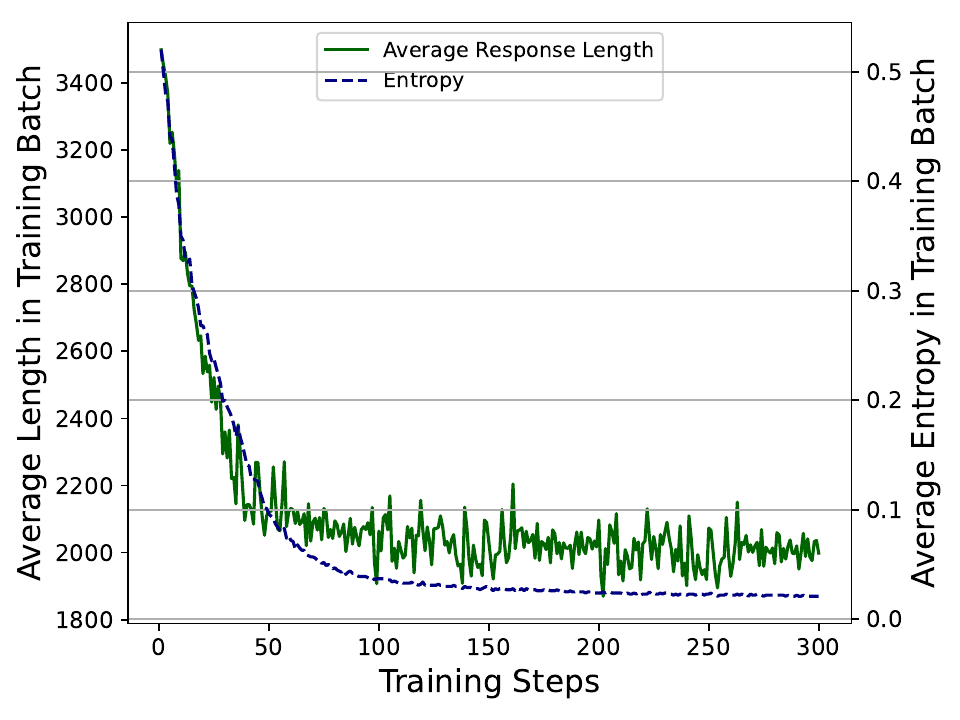}
    \caption{Batch-wise Reward Normalization \newline}
    \label{fig:training_dynamic_a}
  \end{subfigure}
  \hfill
  \begin{subfigure}[b]{0.45\textwidth}
    \centering
    \includegraphics[width=\textwidth]{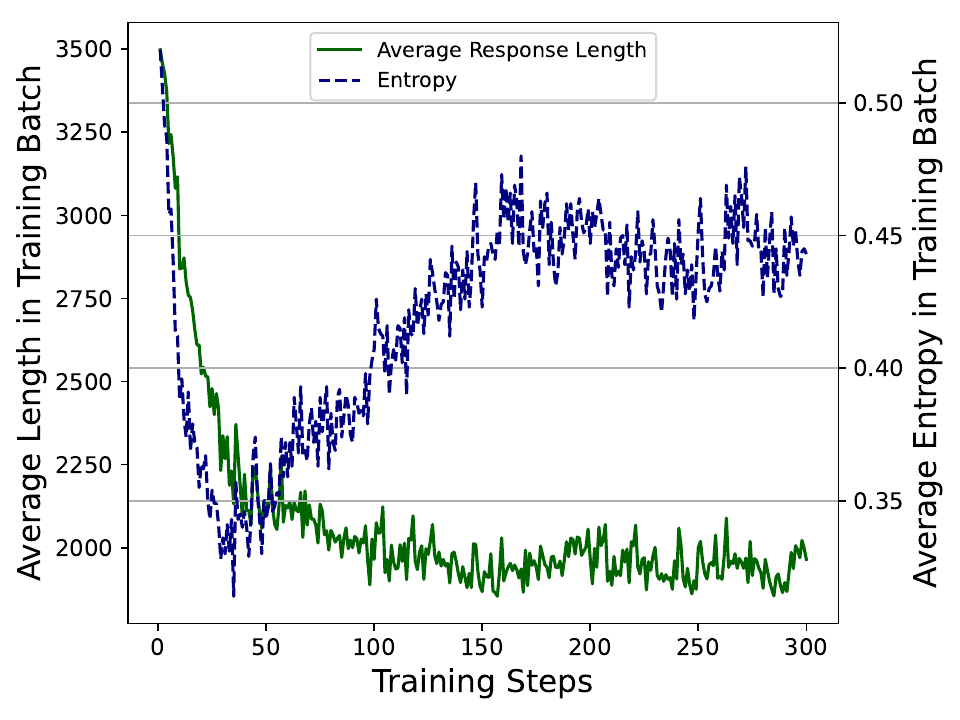}
    \caption{Batch-wise Reward Normalization w/ Higher Clipping Threshold}
    \label{fig:training_dynamic_b}
  \end{subfigure}

  \vspace{0.5em} 

  \begin{subfigure}[b]{0.45\textwidth}
    \centering
    \includegraphics[width=\textwidth]{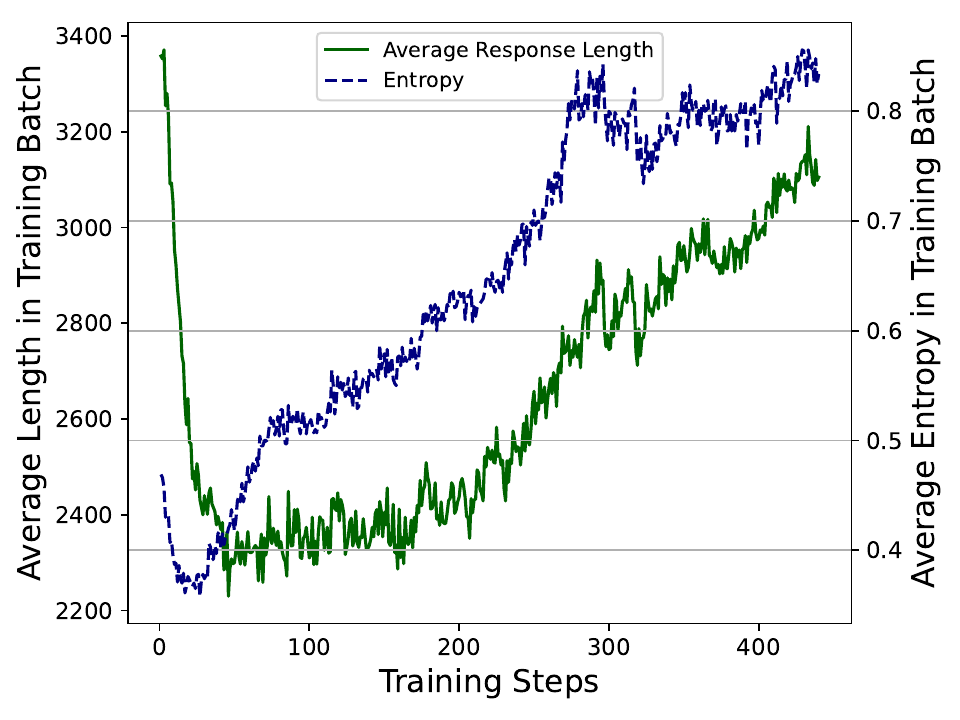}
    \caption{\textbf{DLER}: Batch-wise Reward Normalization w/ Higher Clipping Threshold and w/ Dynamic Sampling}
    \label{fig:training_dynamic_c}
  \end{subfigure}

  \caption{Average token entropy and response length per training batch across three RL runs: batch-wise reward normalization, batch-wise normalization with a higher clipping threshold, and our DLER method. DLER not only resolves entropy collapse but also shows rising entropy and gradually increasing response length after an initial drop, suggesting active exploration under the length penalty, unlike the plateaued behavior of the baselines.}
  \label{fig:training_dynamic}
\end{figure}

Building on our findings, we unify batch-wise reward normalization, a higher policy update clipping threshold, dynamic sampling to remove instances lacking balanced training signals, and a simple length truncation penalty into a comprehensive training recipe, which we term \textbf{DLER} (\textbf{D}oing \textbf{L}ength p\textbf{E}nalty \textbf{R}ight). We apply DLER to train DeepSeek-R1-7B for 450 steps and compare its trajectory of accuracy and response length on AIME-24 (Fig.~\ref{fig:our_aime}) and training dynamics (Fig.~\ref{fig:training_dynamic}). 
All ingredients work complementary to each other. Together, these components systematically address the optimization challenges identified in previous sections and synergistically contribute to DLER's superior performance.

\subsection{Difficulty-Aware DLER}
We present a difficulty-aware extension of DLER (DA-DLER) that improves efficiency by adaptively assigning truncation lengths according to question difficulty, yielding greater redundancy reduction than a fixed truncation scheme. Question difficulty is estimated from the correctness ratio of model responses, and truncation targets are dynamically adjusted across difficulty tiers. Concretely, for a question $q$ with a sampled response set $G$, the correctness ratio is defined as the fraction of correct responses in $G$. Each question is then categorized into one of $n$ difficulty levels $\{d_i\}_{i=1}^n$, each associated with a truncation length $\{l_i\}_{i=1}^{n+1}$. For instance, if the correctness ratio lies between $d_i$ and $d_{i+1}$, the truncation length $l_i$ is applied. DA-DLER pushes the boundary of reasoning efficiency by encouraging the model to solve questions it already handles reliably with even fewer reasoning steps, reducing unnecessary token usage.

\section{Experiment} 

\subsection{Setup}
\label{sec:setup}
We conduct our experiments on enhancing the reasoning efficiency of DeepSeek-R1-1.5B/7B~\cite{guo2025Deepseek}, which are widely used as baseline models by prior work~\cite{chen2025verithinker, liu2025laser, zhang2025adaptthink, aggarwal2025l1, luo2025o1pruner}. Training is performed on the DeepScaleR-Preview-Dataset~\cite{deepscaler2025}, a mathematics dataset containing 40K competition-level problems, using veRL~\cite{sheng2024hybridflow}, a reinforcement learning training library. We adopt the original prompt format from DeepSeek-R1~\cite{guo2025Deepseek}, set the number of rollouts per training prompt to 16, and use a batch size of 512. We set the target length of the truncation length penalty to 4000 tokens. The complete list of training hyperparameters is provided in the Appendix.~\ref{appendix:verl_hyperparameters}. This setup is also applied consistently across all experiments presented in earlier sections. We denote the resulting models as \textbf{DLER-R1-1.5B/7B}, we then further apply the difficulty-aware DLER training recipe on DLER-R1-1.5B and 7B for another 150 steps and denote them as \textbf{DA-DLER-R1-1.5B/7B}. For DA-DLER, the difficulty threshold is set to a correctness ratio of ${0.5}$, with corresponding truncation length penalties of ${2000, 4000}$ tokens.
We compare our models against prior publicly released models trained on DeepSeek-R1-1.5B and 7B, including:

\begin{itemize}
 \item Laser~\cite{liu2025laser} introduces a difficulty-aware length penalty reward, releasing two checkpoints—\textbf{Laser-DE-L4096-1.5B} and \textbf{Laser-DE-L4096-7B}—trained using the same dataset as ours.
 \item AdaptThink~\cite{zhang2025adaptthink} employs reinforcement learning to enable the model to skip the reasoning process for simpler questions. Using the same training dataset as ours, it provides two checkpoints: \textbf{AdaptThink-1.5B-delta0.05} and \textbf{AdaptThink-7B-delta0.05}.
 \item LC-R1~\cite{cheng2025optimizing} proposes a new length penalty reward targeting overall conciseness along with a Compress Reward aimed at removing invalid portions of the reasoning process. Although trained on a different dataset, it releases two checkpoints: \textbf{LCR1-1.5B} and \textbf{LCR1-7B}.
\item VeriThinker~\cite{chen2025verithinker} is an SFT-based approach that fine-tunes the model to improve self-reflection and eliminate redundant reasoning steps, releasing a single checkpoint, \textbf{VeriThinker-7B}.
\end{itemize}
We evaluate DLER models and these baselines on AIME-24~\cite{aime24}, AMC (AMC 2022 and AMC 2023)~\cite{amc}, MATH~\cite{hendrycks2021math500}, Minerva~\cite{lewkowycz2022minerva} and Olympiad Bench~\cite{he2024olympiadbench}. All evaluations are conducted using vLLM as the inference backend with a sampling temperature of 0.6, $top_p$ = 0.95, and a maximum response length of 32k tokens. For each benchmark prompt, we generate 16 samples and compute the average pass@1 score.

\subsection{Main Results}
\label{sec:main_results}

\begin{table}[!htp]\centering
\caption{Comparison of DLER models and baseline models in terms of Pass@1 accuracy and corresponding average output length (tokens) across benchmarks.}
\label{tab:main_table}
\resizebox{\textwidth}{!}{
\begin{tabular}{lcc|cc|cc|cc|cc|cc}\toprule
&MATH $\uparrow$ &Length $\downarrow$&AIME-24 $\uparrow$&Length $\downarrow$&AMC $\uparrow$ &Length $\downarrow$&Minerva $\uparrow$ &Length $\downarrow$&Olympiad $\uparrow$ &Length $\downarrow$&Total Avg $\downarrow$\\\cmidrule{2-12}
DeepSeek-R1-1.5B &84.31 &5500  &29.79 &16916 &61.97 &10967  &38.41 &7494  &44.07 &11620  &10499 \\\cmidrule{1-12}
LCR1-1.5B &81.80 &2612  &21.04 &9335  &59.64 &5377  &40.64 &2702  &41.58 &5876  &5180 \\\cmidrule{1-12}
Laser-DE-L4096-1.5B &85.27 &2685  &30.62 &8194  &68.14 &4890  &42.69 &3322  &46.21 &5323  &4882 \\\cmidrule{1-12}
AdaptThink-1.5B-delta0.05 &82.26 &1651  &30.21 &7550  &61.37 &3622  &41.52 &\textbf{\textcolor{NvidiaGreen}{1745}}  &42.50 &4279  &3769 \\\cmidrule[1pt]{1-12}
\textbf{DLER-R1-1.5B}&\textbf{\textcolor{NvidiaGreen}{86.95}} &1652  &\textbf{\textcolor{NvidiaGreen}{34.38}} &3551  &70.48 &2537  &43.59 &2029  &48.31 &2563  &\textbf{\textcolor{Green}{2466 (-77\%)}} \\\cmidrule{1-12} 
\textbf{DA-DLER-R1-1.5B} & 86.70 & \textbf{\textcolor{NvidiaGreen}{1484}}  & 34.37 & \textbf{\textcolor{NvidiaGreen}{2888}}  & \textbf{\textcolor{NvidiaGreen}{72.36}} & \textbf{\textcolor{NvidiaGreen}{2154}}  & \textbf{\textcolor{NvidiaGreen}{44.89}} & 1895  & \textbf{\textcolor{NvidiaGreen}{48.70}} & \textbf{\textcolor{NvidiaGreen}{2109}}  & \textbf{\textcolor{NvidiaGreen}{2106 (-80\%)}}
\\
\bottomrule
\\
DeepSeek-R1-7B &93.60 &3999  &55.40 &13241  &82.90 &7461  &49.79 &5199  &58.21 &8837  &7747 \\\cmidrule{1-12}
R1-VeriThinker-7B &93.63 &2591  &51.25 &9805  &81.77 &5611  &46.14 &2934  &57.92 &6470  &5482 \\\cmidrule{1-12}
LCR1-7B &90.65 &1534  &50.20 &7305  &79.29 &3609  &50.32 &\textbf{\textcolor{NvidiaGreen}{1559}}  &55.96 &4352  &3671 \\\cmidrule{1-12}
Laser-DE-L4096-7B &93.48 &1759  &55.20 &5691  &82.83 &3262  &50.22 &1884  &57.90 &3451  &3209 \\\cmidrule{1-12}
AdaptThink-7B-delta0.05 &91.38 &2005  &53.12 &10250  &81.47 &5461  &50.67 &2522  &56.96 &6434  &5334\\\cmidrule[1pt]{1-12}
\textbf{DLER-R1-7B} &\textbf{\textcolor{NvidiaGreen}{94.21}} &1634  &\textbf{\textcolor{NvidiaGreen}{55.62}} &3230  &84.41 &2512  &\textbf{\textcolor{NvidiaGreen}{53.88}} &2058  &60.48 &2592  &\textbf{\textcolor{Green}{2405 (-69\%)}}
\\\cmidrule{1-12}
\textbf{DA-DLER-R1-7B} &94.17 & \textbf{\textcolor{NvidiaGreen}{1481}}  & 53.90 & \textbf{\textcolor{NvidiaGreen}{2878}}  &\textbf{\textcolor{NvidiaGreen}{84.56}} & \textbf{\textcolor{NvidiaGreen}{2286}}  &53.60 & 1896 & \textbf{\textcolor{NvidiaGreen}{61.16}} & \textbf{\textcolor{NvidiaGreen}{2296}}  & \textbf{\textcolor{NvidiaGreen}{2167 (-73\%)}}
\\
\bottomrule
\end{tabular}}
\end{table}

Table~\ref{tab:main_table} reports the performance of DLER and DA-DLER compared with prior state-of-the-art reasoning compression baselines across five benchmarks—MATH, AIME-24, AMC, Minerva, and Olympiad—together with average response length. For compressing the reasoning traces of DeepSeek-R1-1.5B, DLER-R1-1.5B achieves the strongest accuracy on all benchmarks while reducing response length to an average of 2466 tokens, over 4× shorter than the original DeepSeek-R1-1.5B and 51–35\% shorter than previous baselines. In particular, it attains 86.95 on MATH, 34.38 on AIME, and 48.31 on Olympiad, surpassing Laser-DE by margins of 1.68, 3.76, and 2.10 points, respectively. Building on DLER-R1-1.5B, DA-DLER-R1-1.5B further improves efficiency by dynamically adapting different truncation lengths. It maintains comparable accuracy to DLER while cutting the average response length by an additional 15\% (from 2466 to 2106), demonstrating that difficulty-aware truncation can further push efficiency beyond fixed-penalty training.

A similar pattern holds for the 7B model. DLER-R1-7B establishes new state-of-the-art results, reaching 94.21 on MATH, 55.62 on AIME, and 84.41 on AMC, while compressing average response length to 2405 tokens—69\% shorter than the original DeepSeek-R1-7B and 25\% shorter than Laser-DE. Importantly, DLER-R1-7B not only preserves but improves accuracy over the base model on all the benchmarks. DA-DLER-R1-7B then pushes efficiency further, reducing average length by an additional 12\% without compromising accuracy.

Overall, DLER models achieves the best trade-off between reasoning accuracy and efficiency across both model sizes. Competing methods either maintain high accuracy at significantly longer response length or reduce token usage at the expense of accuracy. By contrast, DLER enables the model to retain strong reasoning ability while producing drastically shorter responses. The consistent gains across diverse reasoning benchmarks, particularly on challenging datasets such as AIME-24 and Olympiad, further demonstrate the robustness of our optimization recipe.

\subsection{Performance Under Different Test-time Scaling Settings}

\begin{figure}[h]
    \centering
  \begin{subfigure}[b]{0.85\textwidth}
    \centering
    \includegraphics[width=\textwidth]{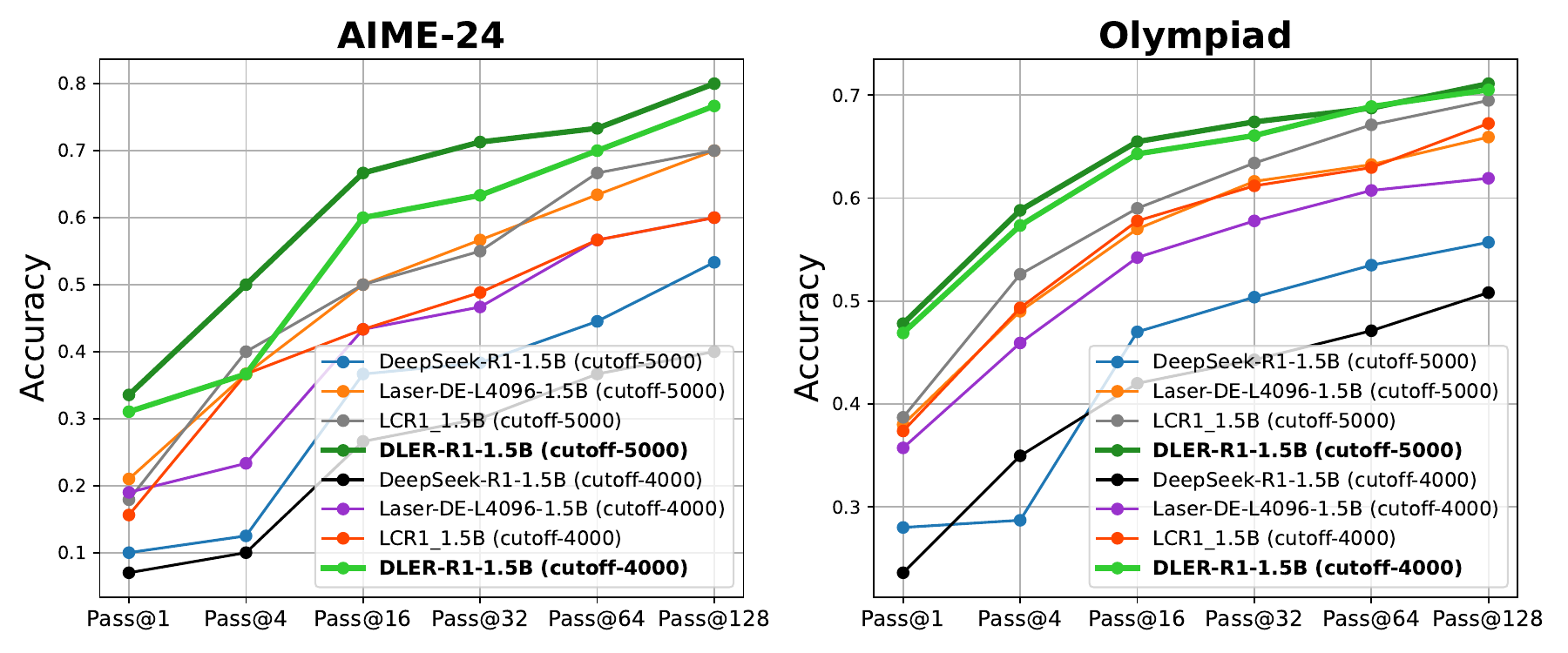}
    \caption{DLER-R1-1.5B vs Baselines}
    \label{fig:sub1}
  \end{subfigure}
  \vskip\baselineskip
  \begin{subfigure}[b]{0.85\textwidth}
    \centering
    \includegraphics[width=\textwidth]{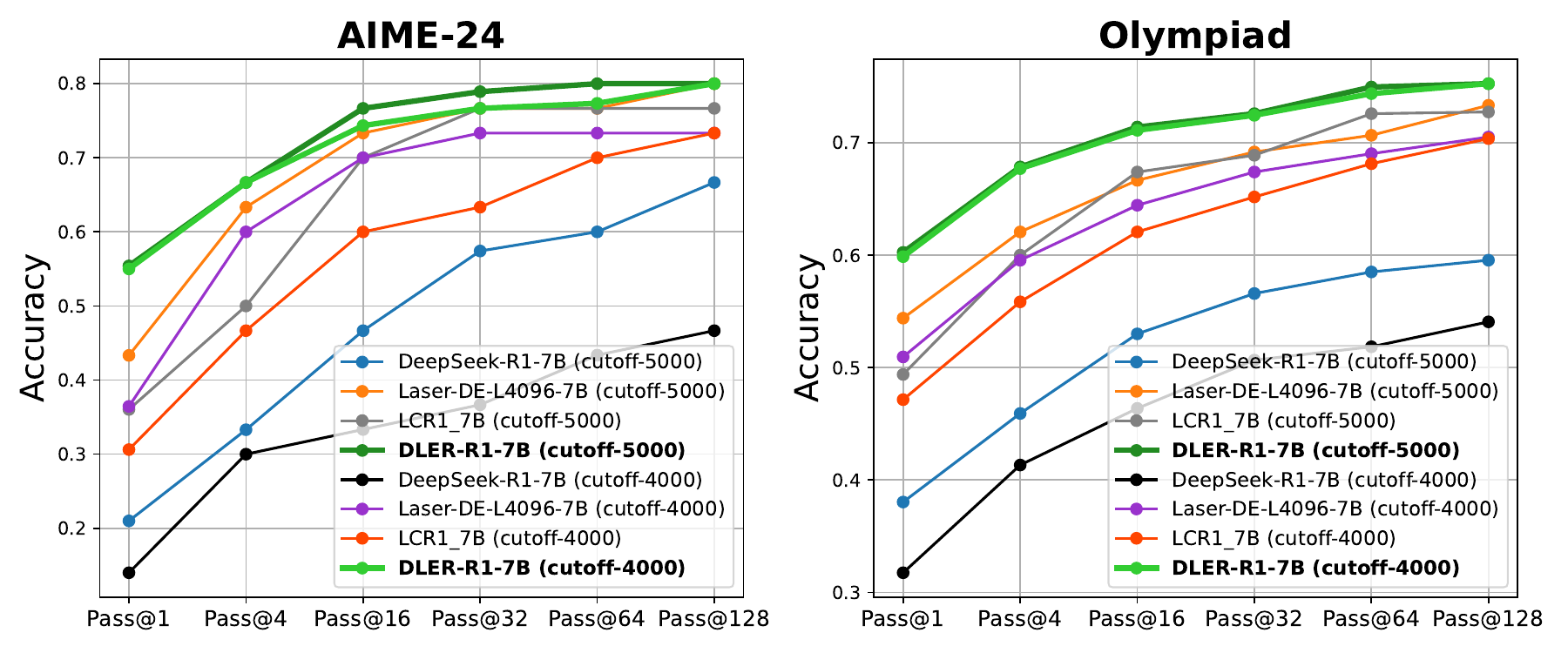}
    \caption{DLER-R1-7B vs Baselines}
    \label{fig:sub2}
  \end{subfigure}
\caption{Pass@K accuracy of DLER-1.5B/7B versus baselines on AIME-24 and Olympiad under different length cutoffs. DLER consistently outperforms all baselines across different test-time budget settings.}
\label{fig:passk}
\end{figure}

We further compare the performance of DLER-R1-1.5B and 7B models against Laser-DE-L4096-1.5B/7B, LCR1\_1.5B/7B and the original DeepSeek-R1 models under different test-time scaling by evaluating Pass@K under various response length cutoffs.

Fig.~\ref{fig:passk} shows Pass@\{1,4,16,32,64,128\} accuracy on two reasoning benchmarks—AIME-24 and Olympiad—for both the 1.5B (top row) and 7B (bottom row) models under hard-cutoffs of 4000 and 5000 tokens. Across all Pass@K settings, DLER-R1-1.5B/7B outperform the baselines, with the margin being most pronounced under the more restrictive token limit. For example, on AIME-24 with cutoff set to 4000, DLER-R1-7B model achieves the highest Pass@1 accuracy and sustains its advantage as Pass@K increases. A similar pattern emerges on the Olympiad benchmark, where our models consistently exceed the performance of Laser-DE and LCR1 baselines. Even under stringent length constraints, our approach preserves high accuracy, confirming its robustness in generating concise yet precise outputs. These results underscore the practical benefits and superiority of DLER models for achieving efficient and scalable reasoning across model sizes and different test-time scaling budget.

\subsection{DLER Enables Superior Test-time Scaling through Parallel Thinking}
\label{sec:parallel_thinking}

In the previous section, we demonstrated that DLER models maintain superior accuracy across different test-time token budgets. Here, we extend the analysis to show that reasoning efficiency also also translates into superior test-time scaling in terms of accuracy/latency. We benchmark parallel thinking latency, defined as the average request (per-question) time required to parallelly generate multiple responses. All experiments are conducted using vLLM on a single NVIDIA H100 GPU with a response length cutoff of 32000 tokens. We evaluate performance on AIME-24, reporting both Pass@K accuracy and the average request time required to generate K responses for each question.

As shown in Fig.~\ref{fig:parallel_latency_teaser} and Table~\ref{tab:parallel_thinking_latency}, DLER-R1-1.5B achieves substantial latency improvements. For single-response inference, average request time drops from 58.99 seconds with DeepSeek-1.5B to just 12.35 seconds, a 4.8× speedup. To reach 80.00 accuracy, DLER-1.5B requires only 52.09 seconds with 128 rollouts, compared to 229.00 seconds for DeepSeek-1.5B with 64 rollouts—a 176.91-second reduction (78\% less time). Strikingly, the average time required by DLER-1.5B to produce 128 rollouts is still lower than that needed by DeepSeek-1.5B to generate a single response, all while delivering nearly 50\% greater accuracy. The same advantage holds at the 7B scale. DLER-R1-7B reduces single-response request time of DeepSeek-R1-7B from 93.43 seconds to 23.73 seconds, nearly a 4× improvement. To achieve 83.33 accuracy, DLER-R1-7B requires only 85.43 seconds with 256 rollouts, while DeepSeek-7B takes 221.22 seconds with 16 rollouts—representing a 135.79-second reduction (62\% less time). Moreover, even with 256 rollouts, DLER-R1-7B is still faster than DeepSeek-7B producing a single response on average, while yielding an additional 30\% accuracy improvement.

\textbf{These results signal a fundamental shift in perspective. Whereas prior work~\cite{guo2025Deepseek,liu2025prorl} has largely emphasized maximizing per-response accuracy through extended reasoning traces, our findings show that prioritizing reasoning efficiency (intelligence per token) yields far greater benefits by enabling superior test-time scalability.} More concretely, it makes more sense to allocate test-time compute to an efficient reasoning model rather than one that may achieve slightly higher Pass@1 accuracy but requires up to 5× more time to match the accuracy of the efficient model once parallel thinking is applied. Overall, DLER-trained models achieve higher accuracy in substantially less wall-clock time, making them a far more practical option for real-world deployment.

\subsection{Different Length Penalties No Longer Push the Accuracy–Efficiency Frontier}
\label{sec:diff_length_peanlties}

\begin{figure}[ht]
  \centering
  \begin{subfigure}[b]{0.45\textwidth}
    \centering
    \includegraphics[width=\textwidth]{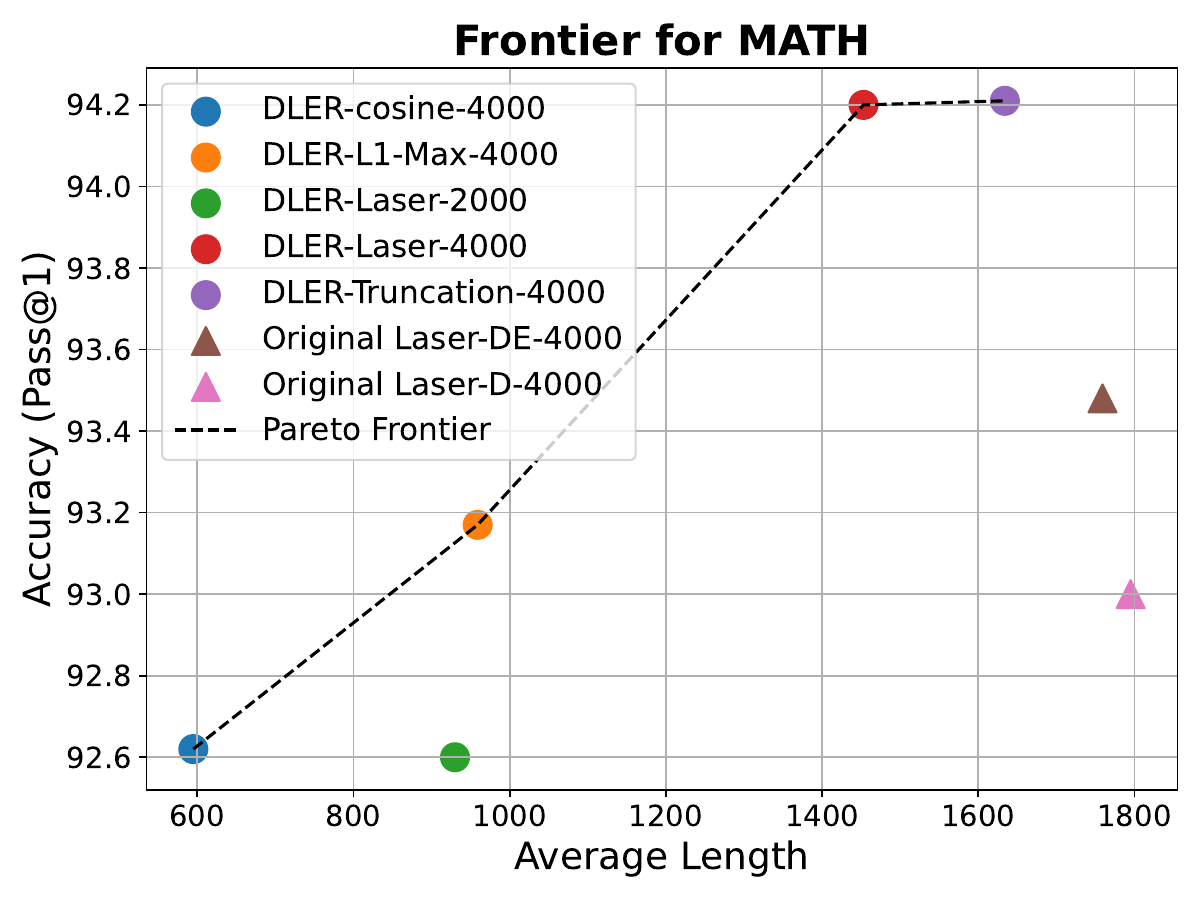}
    \caption{MATH}
    \label{fig:diff_length_penalties_math}
  \end{subfigure}
  \hfill
  \begin{subfigure}[b]{0.45\textwidth}
    \centering
    \includegraphics[width=\textwidth]{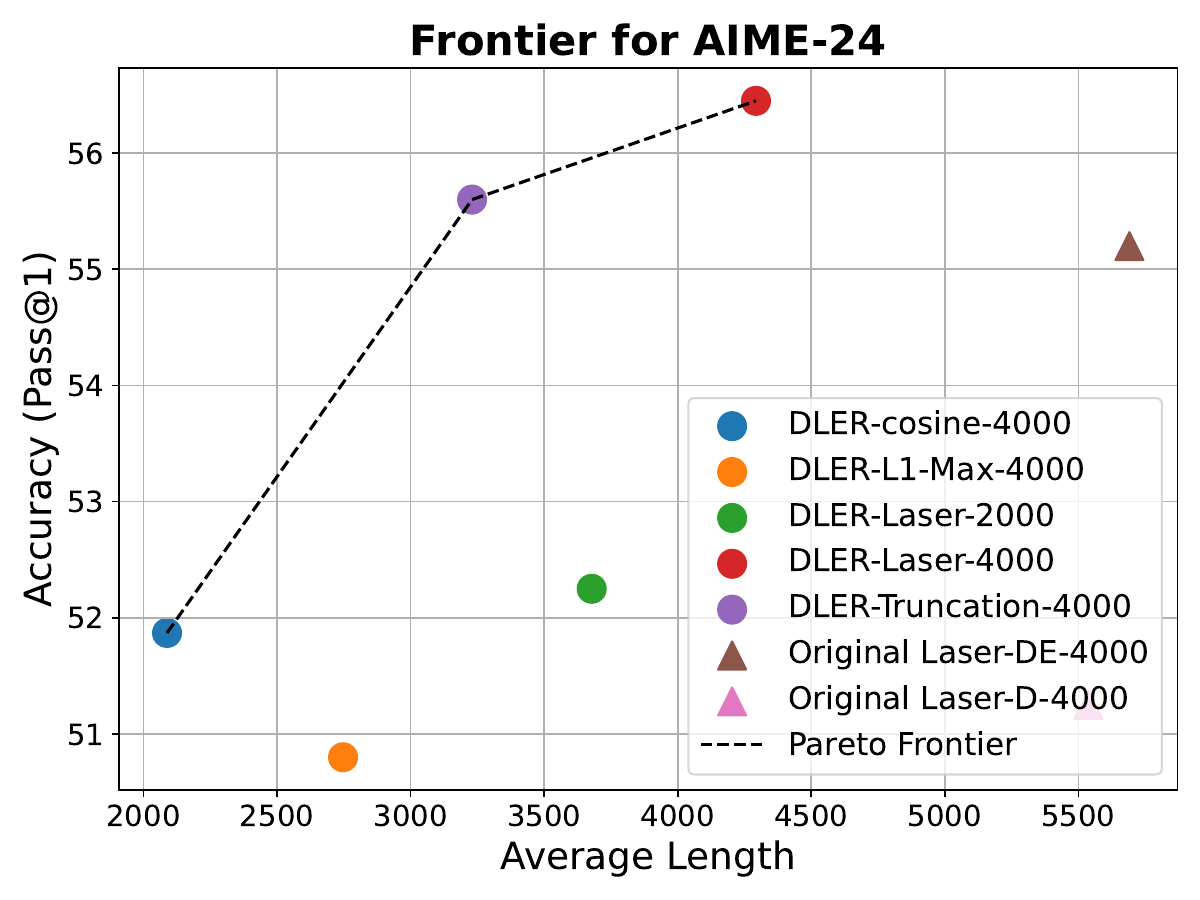}
    \caption{AIME-24}
    \label{fig:diff_length_penalties_aime}
  \end{subfigure}
  \vspace{0.5em} 
  \begin{subfigure}[b]{0.45\textwidth}
    \centering
    \includegraphics[width=\textwidth]{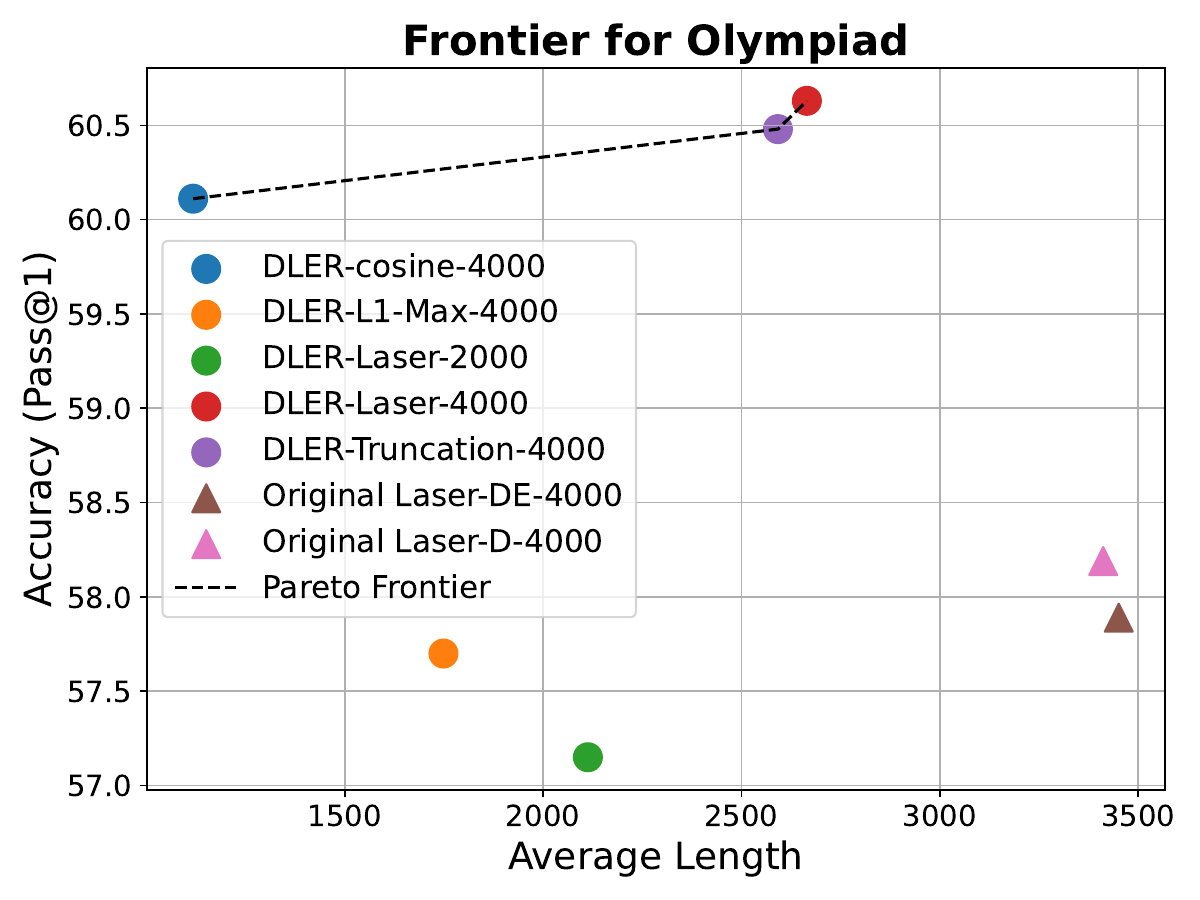}
    \caption{Olympiad}
    \label{fig:diff_length_penalties_olympiad}
  \end{subfigure}

  \caption{Accuracy and average response length of DeepSeek-R1-7B trained using DLER with different length penalties on MATH, AIME-24, and Olympiad. DLER establishes a new accuracy–length efficiency frontier, with varying length penalties moving performance along the frontier rather than beyond it.}
  \label{fig:diff_length_penalties}
\end{figure}

In this section, we show that with our proposed optimization recipe (DLER), the effect of adopting different length-penalty rewards fundamentally changes. Specifically, the accuracy–length relationship is no longer altered in a way that yields strictly shorter responses with higher accuracy; instead, a trade-off always exists. We illustrate this by adopting several different length penalties from prior works with DLER: \textit{Truncation}—our default penalty—assigns zero reward when exceeding a target length; \textit{Cosine} scales the penalty according to the cosine of the deviation from the target length; \textit{L1-Max} \cite{aggarwal2025l1}; and \textit{Laser} \cite{liu2025laser}. Here, “DLER-xxx-4000” denotes training with DLER using length penalty xxx and a target length of 4000. Results for the original Laser-DE-4000 and Laser-D-4000 are based on the publicly released models from \cite{liu2025laser}.

Fig.~\ref{fig:diff_length_penalties} shows that across all benchmarks, models trained with DLER consistently outperform their non-DLER counterparts. In particular, Laser-4000 achieves higher accuracy than both Original Laser-DE-4000 and Original Laser-D-4000 while attaining shorter or comparable average lengths, as shown for MATH (Fig.~\ref{fig:diff_length_penalties_math}), AIME-24 (Fig.~\ref{fig:diff_length_penalties_aime}), and Olympiad (Fig.~\ref{fig:diff_length_penalties_olympiad}). Furthermore, in all three tasks, the accuracy/average length Pareto frontiers are defined entirely by models trained with DLER, establishing a new optimal boundary in the accuracy–length space. Under this setting, varying the length-penalty reward does not push performance beyond the frontier; rather, it shifts the point along it.

Notably, the simplest length penalty—truncation—remains highly effective compared to other complicated length penalty functions, consistently producing frontier points competitive with or superior to more complex penalties such as L1-Max and Laser. Importantly, truncation requires significantly less training time because it terminates rollouts upon reaching the targeted cutoff length, whereas L1-Max and Laser operate without a hard length cutoff and therefore still require full-length rollouts. This makes truncation not only a strong accuracy–length trade-off option but also the most computationally efficient choice in terms of training cost.

\subsection{Overcoming Quality Limitations of Publicly Available Training Data}
\label{sec:weight_merging}

\begin{table}[h]\centering
\caption{Comparison of Llama-3.1-Nemotron-Nano-8B-v1 (Nemotron-8B), DLER-Llama-Nemotron-8B (DLER-Nemotron-8B) and the merged model \textbf{DLER-Nemotron-8B-Merge} (DLER-Nemotron-8B-Merge) in terms of Pass@1 accuracy and corresponding average output length (tokens) across benchmarks.}
\label{tab:model_merging}
\resizebox{ \textwidth}{!}{
\begin{tabular}{lcc|cc|cc|cc|cc|c}
\toprule
                        Model &  MATH $\uparrow$ &  Length $\downarrow$ &  AIME-24 $\uparrow$& Length $\downarrow$ &   AMC $\uparrow$ &  Length $\downarrow$&  Minerva $\uparrow$ &Length $\downarrow$ &  Olympiad $\uparrow$ &Length $\downarrow$ &  Total Avg $\downarrow$ \\
\midrule
Nemotron-8B &     95.40 &        3069 & 66.40 &           9899 & 88.25 &          6228 &    52.38 &              4031 &     64.33 &               6755 &       5996 \\\cmidrule{1-12}
                DLER-Nemotron-8B &     95.00 & \textbf{\textcolor{NvidiaGreen}{1843}} & 63.54 & \textbf{\textcolor{NvidiaGreen}{3867}} & 88.47 & \textbf{\textcolor{NvidiaGreen}{2850}} & \textbf{\textcolor{NvidiaGreen}{54.27}} & \textbf{\textcolor{NvidiaGreen}{2276}} & \textbf{\textcolor{NvidiaGreen}{65.63}} & \textbf{\textcolor{NvidiaGreen}{2843}} & \textbf{\textcolor{NvidiaGreen}{2735 (-55\%)}} \\
      \textbf{DLER-Nemotron-8B-Merge} &     \textbf{\textcolor{NvidiaGreen}{95.20}} & 1995  & \textbf{\textcolor{NvidiaGreen}{66.66}} & 5013 & \textbf{\textcolor{NvidiaGreen}{89.23}} & 3358 & 53.19 & 2301 & 65.39 & 3520 &    \textbf{\textcolor{LimeGreen}{3237 (-46\%)}} \\
\bottomrule
\end{tabular}}
\end{table}

Although reasoning models are evolving at an accelerated pace, the accompanying training datasets are rarely made public. Consequently, practitioners are constrained to employ publicly available datasets, whose difficulty often falls short of matching the capacity of state-of-the-art models. To investigate this potential problem, we apply DLER—incorporating a truncation length penalty with a target length of 6000 tokens and a higher clipping threshold threshold of 0.36—to Llama-3.1-Nemotron-Nano-8B-v1, a high-capacity baseline that outperforms DeepSeek-32B on MATH and matches DeepSeek-14B on AIME-24. As shown in Table~\ref{tab:model_merging}, the fine-tuned model, DLER-Nemotron-8B, achieves a substantial reduction in average response length, from 6728 to 2735 tokens (-55\%), with reductions observed across all evaluated benchmarks. Despite these efficiency gains, we observe accuracy degradation on MATH (95.40 to 95) and AIME-24 (66.40 to 63.54), while AMC, Minerva, and Olympiad exhibit slight improvements.

To fully recover the accuracy degradation without access to better proprietary training data, we explore model merging as a complementary approach. Prior work \cite{mukherjee2025reinforcement} has shown that reinforcement learning fine-tuning produces relatively small and sparse parameter updates across weight matrices. This suggests that merging the original model with our efficient fine-tuned variant could retain efficiency gains while recovering lost accuracy, as the resulting weight deltas are modest and thus amenable to merging. However, our initial experiments with naive strategies—such as parameter averaging or linear interpolation—failed to strike the desired trade-off, yielding either minimal accuracy recovery or a substantial increase in response length.
We therefore adopt a update-selective merging approach, inspired by \cite{yadav2023ties}, where we retain only the top 25\% of largest-update parameter deltas from the efficient model, then scale by a factor of 0.7, and add to the original model parameters. The resulting merged model (DLER-Nemotron-8B-Merge) restores the lost accuracy on MATH (95.20, -0.20 vs base) and AIME-24(66.66, +0.26 vs base), while further improving AMC and maintaining competitive Minerva and Olympiad accuracy. Importantly, it sustains a substantial reduction in average response length ( -46\%), preserving the efficiency improvements from DLER training. In summary, when fine-tuning high-capacity reasoning models on public datasets results in accuracy degradation, update-selective weight merging technique provides an effective remedy. This approach enables near-complete recovery of baseline accuracy while retaining large reductions in response length, offering a practical and training-free pathway to producing both accurate and efficient reasoning models.

\subsection{Analysis}

\subsubsection{Entropy Distribution}

\begin{figure}[h]
  \centering
  \begin{subfigure}[b]{0.3\textwidth}
    \centering
    \includegraphics[width=\textwidth]{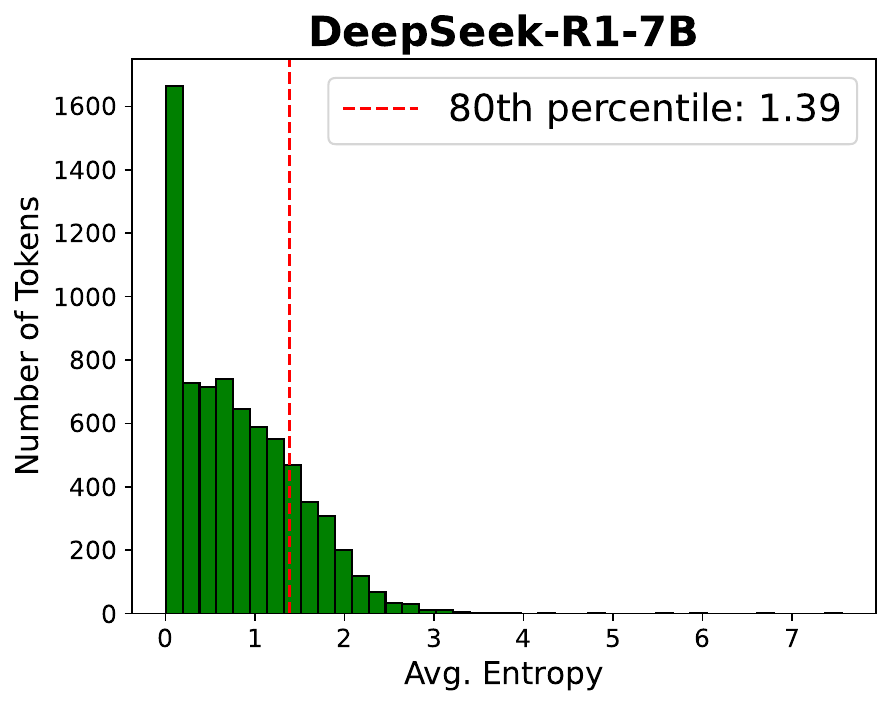}
    \caption{DeepSeek-R1-7B}
    \label{fig:sub1}
  \end{subfigure}
  \begin{subfigure}[b]{0.3\textwidth}
    \centering
    \includegraphics[width=\textwidth]{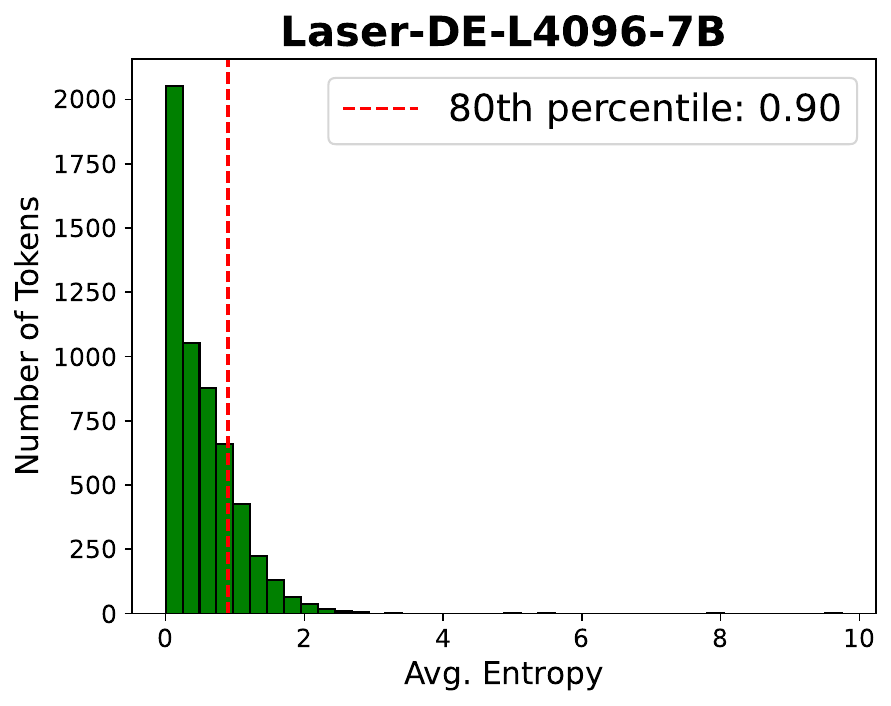}
    \caption{Laser-DE-L4096-7B}
    \label{fig:sub2}
  \end{subfigure}
  \begin{subfigure}[b]{0.3\textwidth}
    \centering
    \includegraphics[width=\textwidth]{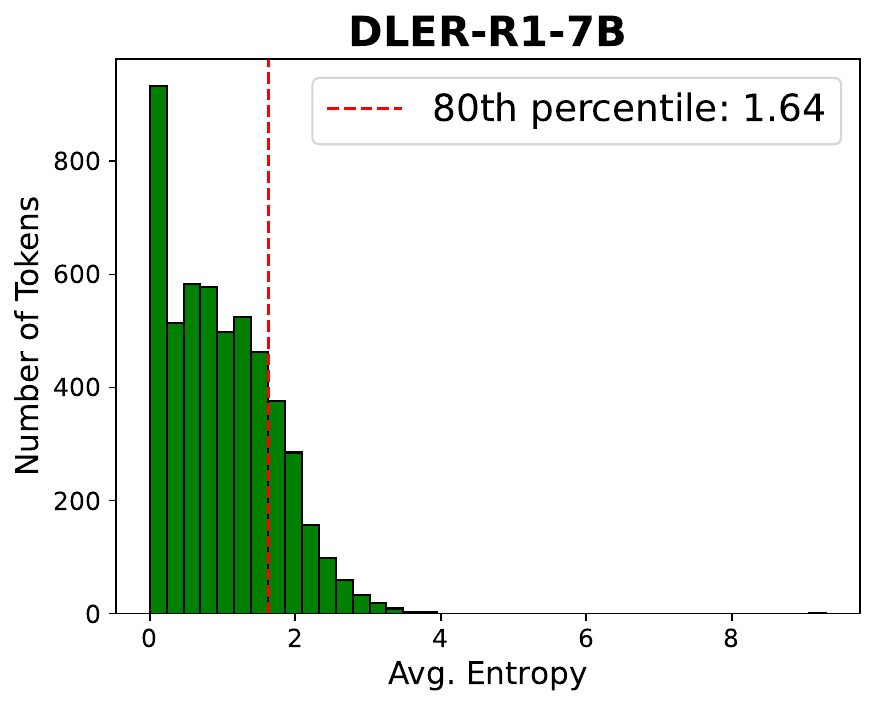}
    \caption{DLER-R1-7B}
    \label{fig:sub3}
  \end{subfigure}

  \caption{Token entropy distribution of DeepSeek-R1-7B, Laser-DE-L4096-7B, and DLER-R1-7B on AIME-24. Laser-DE-L4096-7B shows a markedly contracted distribution relative to DeepSeek-R1-7B, indicating fewer high-entropy tokens and reduced reasoning exploration capability, while DLER-R1-7B exhibits a slight increase in such tokens.}
  \label{fig:distribution}
\end{figure}

In this section, we follow the setup of \cite{wang2025beyond} to examine the distribution of generation entropy in chain-of-thought reasoning at the token level to evaluate how reasoning compression affects model diversity and exploration ability. We evaluate three models: the original DeepSeek-R1-7B, the released Laser-DE-L4096-7B from \cite{liu2025laser}, and our DLER-R1-7B, generating responses for AIME-24 questions with 16 rollouts per question. Token-level entropy is computed following the procedure in \cite{wang2025beyond}. Across all three models, regardless of whether they have undergone efficient reasoning RL training, we observe a consistent pattern: only a small fraction of tokens exhibit relatively high entropy, while the majority have low entropy. This results in a right-skewed token entropy distribution for all models, consistent with the findings in \cite{wang2025beyond}. However, we also find notable differences post the efficient RL training. The entropy distribution of Laser-DE-L4096-7B contracts significantly, indicating a reduction in the number of high-entropy tokens, whereas DLER-R1-7B shows a slight increase in such tokens. As noted in \cite{wang2025beyond} and our analysis in Sec.~\ref{sec:entropy_collapse}, high-entropy tokens are typically those that initiate exploration and reasoning steps. Therefore, the observed decrease in high-entropy tokens for Laser-DE-L4096-7B suggests diminished exploration capacity, while the modest increase in DLER-R1-7B indicates that our training recipe better preserves this ability. This preservation of exploration contributes to DLER-R1-7B’s superior accuracy and shorter response lengths after RL training.

\subsubsection{Reasoning Trace Analysis}

\begin{table}[!h]\centering \caption{Average tokens per step, total steps, and count of transition keywords per response for DeepSeek-R1-7B, Laser-DE-L4096-7B, and DLER-R1-7B on the AIME-24.}
\label{tab:reasoning_trace_analysis} 
\resizebox{ \textwidth}{!}{ 
\begin{tabular}{lcccccccccc}\toprule &\multicolumn{3}{c}{\#Tokens per step} &\multicolumn{3}{c}{\#Steps} &\multicolumn{3}{c}{\#Keywords} \\\cmidrule{2-10} Model &Overall &Correct &Incorrect &Overall &Correct &Incorrect &Overall &Correct &Incorrect \\\cmidrule{1-10} DeepSeek-R1-7B &34 &34 &34 &461 &245 &736 &207 &85 &361 \\\cmidrule{1-10} Laser-DE-L4096-7B &37 &35 &38 &175 &122 &240 &82 &35 &140 \\\cmidrule{1-10} \textbf{DLER-R1-7B} &29 &29 &30 &118 &108 &\textbf{131 (-83\%)} &51 &28 &\textbf{81 (-78\%)} \\\midrule \bottomrule \end{tabular}} \end{table}

In this section, we analyze the reasoning trajectory to assess the impact of our proposed training recipe on mitigating the overthinking problem in reasoning models. Using the same generation setup on AIME-24 as in the previous entropy distribution analysis, reasoning steps are segmented by double newline ($\textbackslash n \textbackslash n$) delimiters in the model’s output. Following the definition of reasoning keywords in \cite{Lu2025RetroSearchEU}, we designate the following terms as keywords: \{‘But’, ‘Wait’, ‘Alternatively’, ‘However’, ‘Hmm’, ‘Hmmm’, ‘Not sure’, ‘Going back’, ‘Backtrack’, ‘Trace back’, ‘Another’\}. 

Table~\ref{tab:reasoning_trace_analysis} reports reasoning trace statistics on AIME-24 for DeepSeek-R1-7B, Laser-DE-L4096-7B, and DLER-R1-7B. DLER-R1-7B shows a marked reduction in average reasoning steps compared to both baselines, with the most notable improvement in incorrect responses—achieving the usage of only 131 reasoning steps, a 45\% decrease relative to Laser-DE and an 83\% decrease relative to the original model. A similar trend is observed for reasoning keywords, where DLER-R1-7B yields the fewest overall and the lowest number in incorrect cases. As prior works~\cite{chen2025verithinker,Lu2025RetroSearchEU,liu2025laser} have discovered, overthinking is a prevalent phenomenon, particularly when facing challenging questions or uncertainty, often leading to unnecessarily prolonged reasoning or or even near-infinite loops when no final answer is produced. The observed reductions in reasoning steps and keywords demonstrate that our method effectively curtails overthinking, resulting in more concise reasoning trajectories and thereby improving both efficiency and accuracy.

\section{Related Work}
Large reasoning models have demonstrated remarkable capabilities, and promoting their efficient reasoning has become a widely studied topic.
One line of work addresses this by directly modifying the prompts fed to the models. For example, \cite{ma2025nothink} argues that explicit reasoning is not always necessary in low-budget settings and proposes bypassing the reasoning process through simple prompting—directly generating the solution after a prefilled dummy reasoning box.

Another line of work focuses on supervised fine-tuning to improve the efficiency of reasoning models. \cite{Lu2025RetroSearchEU} developed an MCTS-inspired search algorithm to distill concise reasoning traces from large models for fine-tuning student models to improve inference efficiency.
\cite{chen2025verithinker} fine-tune models with auxiliary verification to assess solution correctness, enabling them to determine when self-reflection is necessary and suppress overthinking.
\cite{liu2024canskip} encourage step-skipping by fine-tuning models on self-generated shorter reasoning paths mixed with full-step paths.
\cite{xia2025tokenskip} create fine-tuning datasets of compressed chain-of-thought (CoT) by pruning unimportant tokens from LLM trajectories to automatically trim redundant tokens during reasoning, 
while \cite{ma2025cotvalve} construct such datasets by identifying a direction in the parameter space that can be manipulated to effectively control the length of generated CoT.

A more recent line of effort uses reinforcement learning to promote reasoning efficiency.
\cite{fang2025thinkless} use RL with two control tokens—<short> for concise responses and <think> for detailed reasoning—via Decoupled GRPO, which separately optimizes effective mode selection and response accuracy.
\cite{liu2025laser} present a unified framework that formulates various efficient reasoning methods through the lens of length-based reward shaping, and extend the truncation approach into a novel reward shaping method that uses a step function guided by a desired target length. 
\cite{luo2025o1pruner} first estimate an LLM’s baseline performance through pre-sampling, then apply RL-style fine-tuning to encourage the model to generate shorter reasoning processes while maintaining accuracy.
\cite{hou2025thinkprune} trains long-thinking LLMs via RL with token limits that discard unfinished thoughts and answers beyond the limit for zero reward, with multiple RL rounds using increasingly stringent limits.
\cite{aggarwal2025l1} introduces a simple RL method that optimizes both the correctness of the final output and the generation of reasoning sequences that meet a specified length constraint.

\section{Conclusion}
In this work, we revisited the fundamental problem of improving reasoning efficiency in large reasoning models by re-examining the simplest length penalty—truncation. Our analysis revealed that the limitations observed in prior reinforcement learning approaches stem not from the length penalty design itself, but from instability in the underlying optimization process. By combining batch-wise reward normalization, higher clipping threshold, Dynamic Sampling, and simple length truncation penalty, we introduce the \textbf{Doing Length pEnalty Right (DLER)} recipe. This approach delivers state-of-the-art accuracy-to-length efficiency, fully restoring—and in some cases surpassing—accuracy while cutting response length by more than 70\%. We also propose DA-DLER, a difficulty-aware extension of DLER that dynamically adjusts truncation length based on question difficulty, encouraging the model to shorten responses on easier problems and further improve the efficiency of DLER. We additionally introduce a magnitude-selective weight merging strategy alongside DLER to address the minor accuracy drop observed when applying DLER with public data to high-capacity reasoning models, which are often post-trained on more challenging proprietary corpora. This approach restores nearly all lost accuracy while halving output length, offering a practical, training-free solution for practitioners without access to such proprietary data.

Overall, our findings suggest that improving reasoning efficiency depends more on optimization strategies than on complex penalty designs. This perspective points toward new directions for developing reasoning models that are more accurate, efficient, and accessible, thereby facilitating better test-time scaling.

\newpage
{
  \small
  \bibliographystyle{unsrt}
  \bibliography{main_tech_report}
}

\newpage
\appendix

\section{DLER achieves SOTA Accuracy/Length of CoT trade-off and enable better test-time scaling}

\begin{figure}[h]
\begin{center}
\centering
\begin{subfigure}{0.48\textwidth}
  \centering
  \includegraphics[width=\linewidth]{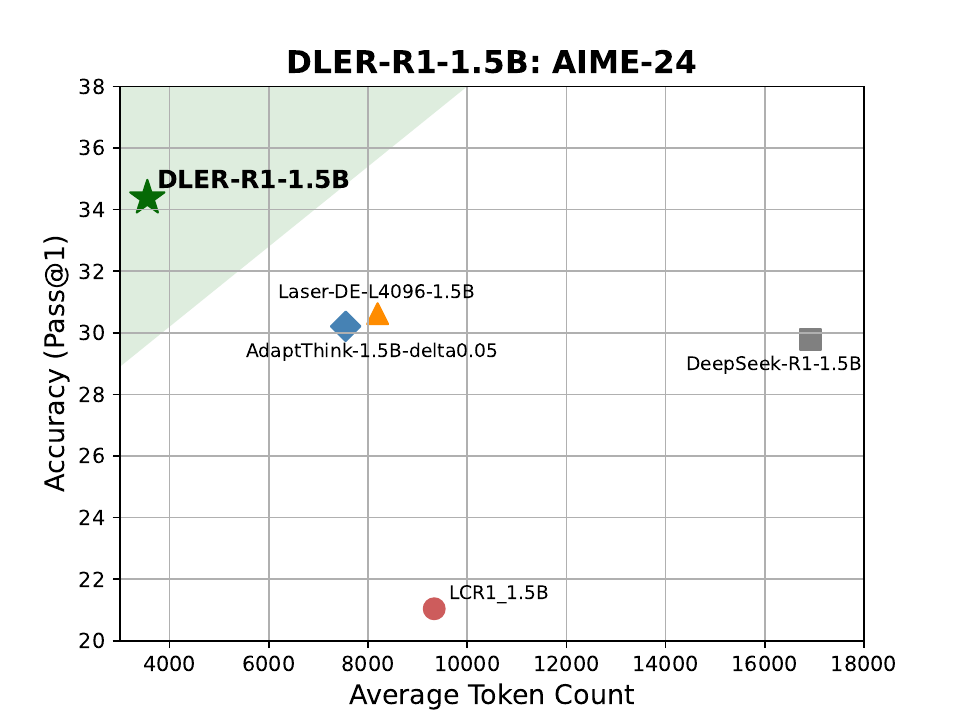}
  
  \caption{DLER Training on DeepSeek-R1-1.5B}
\end{subfigure}
\hfill
\begin{subfigure}{0.48\textwidth}
  \centering
  \includegraphics[width=\linewidth]{figs/plot_TRI_7B.pdf}
  \caption{DLER Training on DeepSeek-R1-7B}
\end{subfigure}
\end{center}
\caption{DLER achieves state-of-the-art Accuracy/Length of CoT trade-off. Compared to baseline models, DLER shortens the CoT by up to $\thicksim$70\% while maintaining accuracy.} 
\label{fig:teaser_appendix}
\end{figure}

\begin{figure}[h]
\begin{center}
\centering
\begin{subfigure}{0.48\textwidth}
  \centering
  \includegraphics[width=\linewidth]{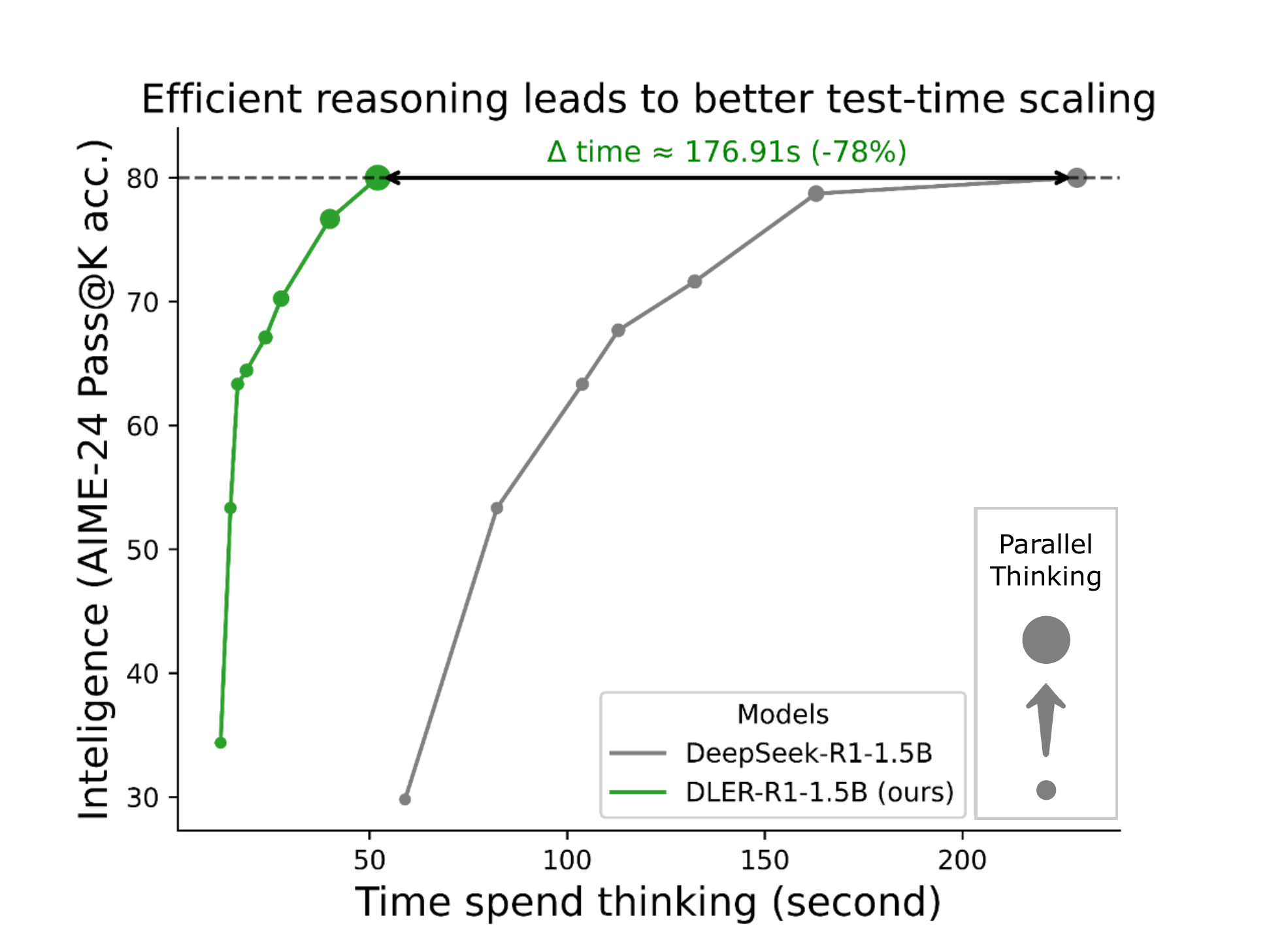}
  
  \caption{DLER-R1-1.5B vs DeepSeek-R1-1.5B}
\end{subfigure}
\hfill
\begin{subfigure}{0.48\textwidth}
  \centering
  \includegraphics[width=\linewidth]{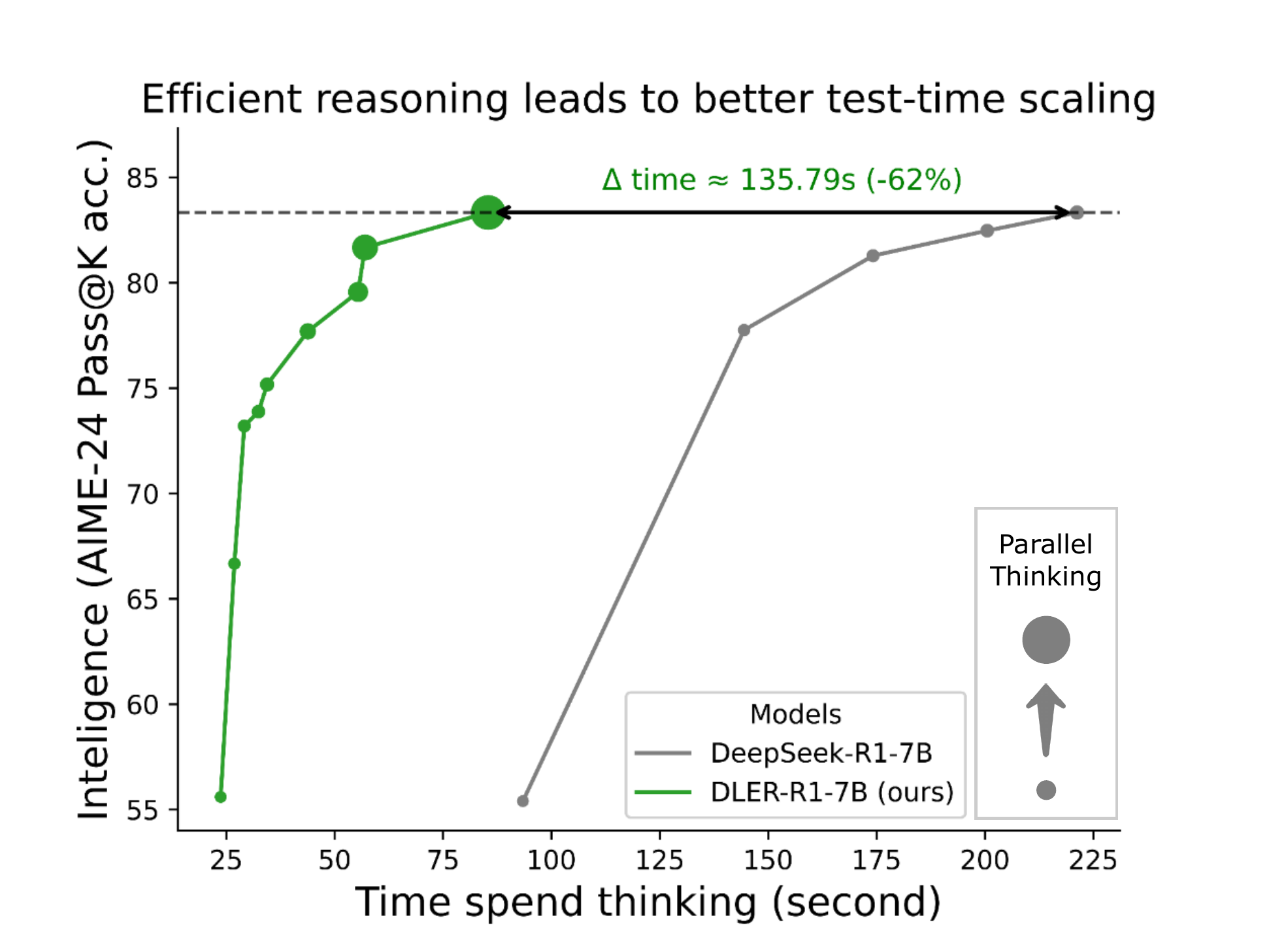}
  \caption{DLER-R1-7B vs DeepSeek-R1-7B}
\end{subfigure}
\end{center}
\caption{
We test the test-time scaling on AIME-24 by varying the number of parallel rollouts (1-256) for DLER-R1 and DeepSeek-R1 models, using vLLM for benchmarking overall latency.
DLER-R1 models demonstrate superior test-time scaling curves compared to DeepSeek-R1 models due to their improved concise reasoning ability.  
} 
\label{fig:parallel_latency_teaser_appendix}
\end{figure}

\newpage
\section{Larger reward variance results in larger bias in advantage estimation.}
\label{appendix:larger_bias}
\subsection*{A.1 Assumptions and Settings}

We observe $N$ rewards $r_i$ for a prompt, and assume the true baseline is $\theta$, such that
\[
r_i = \theta + \epsilon_i, \quad \epsilon_i \sim \mathcal{N}(0, \sigma^2), \quad i = 1, \ldots, N,
\]
with all advantage values $\epsilon_i$ independent. Define
\[
\bar{\epsilon} = \frac{1}{N} \sum_{j=1}^N \epsilon_j, 
\quad 
D = \sqrt{\frac{1}{N} \sum_{j=1}^N (\epsilon_j - \bar{\epsilon})^2}, 
\quad 
A_i = \frac{\epsilon_i - \bar{\epsilon}}{D}.
\]
We will first show that for any finite $N \geq 2$, the advantage estimator $A_i$ is biased:
\[
\mathbb{E}[A_i \mid \epsilon_i] \neq \epsilon_i.
\]
then we will show that given two different $\epsilon_i \sim \mathcal{N}(0, \sigma^2)$ and $\epsilon_i' \sim \mathcal{N}(0, \sigma'^2)$, if $\sigma' > \sigma$ then $\text{Bias}(\epsilon_i') >\text{Bias}(\epsilon_i)$ where $\text{Bias}(\epsilon_i) =
\mathbb{E}[A_i \mid \epsilon_i]$.

Let us first derive why the advantage estimator $A_i$ is biased:

\textbf{Step 1: Bias in the Numerator.}  
The numerator can be expressed as
\[
\epsilon_i - \bar{\epsilon} = \left(1 - \frac{1}{N}\right)\epsilon_i - \frac{1}{N}\sum_{j \neq i}\epsilon_j.
\]
Since the $\epsilon_j$ with $j \neq i$ are zero-mean and independent of $\epsilon_i$, it follows that
\[
\mathbb{E}[\epsilon_i - \bar{\epsilon} \mid \epsilon_i] = \left(1 - \frac{1}{N}\right)\epsilon_i.
\]

\textbf{Step 2: Dependence of the Denominator on $\epsilon_i$.}

(a) Computing $\mathbb{E}[D^2 \mid \epsilon_i]$.  
By definition,
\[
D^2 = \frac{1}{N}\sum_{j=1}^N (\epsilon_j - \bar{\epsilon})^2 
= \frac{1}{N}\sum_{j=1}^N \epsilon_j^2 - \bar{\epsilon}^2.
\]
Because
\[
\bar{\epsilon} = \frac{1}{N}\Big(\epsilon_i + \sum_{j \neq i}\epsilon_j\Big),
\]
and conditioning on $\epsilon_i$ leaves the $\epsilon_j$ (for $j \neq i$) as i.i.d.\ $\mathcal{N}(0,\sigma^2)$, we obtain
\[
\mathbb{E}\!\left[\sum_{j=1}^N \epsilon_j^2 \mid \epsilon_i\right] = \epsilon_i^2 + (N-1)\sigma^2,
\]
\[
\mathbb{E}[\bar{\epsilon}^2 \mid \epsilon_i] = \frac{1}{N^2}\big(\epsilon_i^2 + (N-1)\sigma^2\big).
\]
Thus,
\[
\mathbb{E}[D^2 \mid \epsilon_i] 
= \frac{1}{N}\big(\epsilon_i^2 + (N-1)\sigma^2\big) - \frac{\epsilon_i^2 + (N-1)\sigma^2}{N^2}
= \alpha + \beta \epsilon_i^2,
\]
where $\alpha = \frac{(N-1)^2}{N^2}\sigma^2$ and $\beta = \frac{N-1}{N^2}$.

(b) Non-constancy of $g(\epsilon_i)$.  
Define
\[
g(\epsilon_i) = \mathbb{E}\!\left[\tfrac{1}{D} \,\Big|\, \epsilon_i\right],
\qquad 
\mu(\epsilon_i) = \mathbb{E}[D^2 \mid \epsilon_i] = \alpha + \beta \epsilon_i^2.
\]
Applying a Taylor expansion of $f(x) = x^{-1/2}$ around $x_0 = \mu(\epsilon_i)$ gives
\[
f(x) \approx \frac{1}{\sqrt{x_0}} - \frac{1}{2}\frac{(x-x_0)}{x_0^{3/2}}
+ \frac{3}{8}\frac{(x-x_0)^2}{x_0^{5/2}} + O((x-x_0)^3).
\]
Taking conditional expectation, we obtain
\[
g(\epsilon_i) = \frac{1}{\sqrt{\mu(\epsilon_i)}} 
+ \frac{3}{8}\frac{\mathrm{Var}(D^2 \mid \epsilon_i)}{\mu(\epsilon_i)^{5/2}}.
\]
Since $\mu(\epsilon_i) = \alpha + \beta\epsilon_i^2$ with $\beta > 0$, $g(\epsilon_i)$ depends on $\epsilon_i^2$ and is therefore not constant.

\textbf{Step 3: Combining Results.}  
The estimator can be decomposed as
\[
A_i = \frac{\epsilon_i - \bar{\epsilon}}{D} 
= \left(1 - \frac{1}{N}\right)\frac{\epsilon_i}{D} 
- \frac{1}{N}\Big(\sum_{j \neq i}\epsilon_j\Big)\frac{1}{D}.
\]
For fixed $\epsilon_i$, the distribution of $\sum_{j \neq i}\epsilon_j$ is symmetric about zero, while $1/D$ is always positive. Hence
\[
\mathbb{E}\!\left[-\tfrac{1}{N}\sum_{j \neq i}\epsilon_j \cdot \tfrac{1}{D} \,\Big|\, \epsilon_i\right] = 0.
\]
It follows that
\[
\mathbb{E}[A_i \mid \epsilon_i] = \left(1 - \tfrac{1}{N}\right)\epsilon_i \cdot g(\epsilon_i).
\]

\textbf{Step 4: Concluding the Bias.}  
If $A_i$ were unbiased, we would require
\[
\left(1 - \tfrac{1}{N}\right) g(\epsilon_i) \equiv 1 
\quad \Rightarrow \quad g(\epsilon_i) \equiv \frac{N}{N-1}.
\]
This contradicts Step~2, which showed $g(\epsilon_i)$ depends on $\epsilon_i^2$.  
Therefore, for any finite $N \geq 2$,
\[
\mathbb{E}[A_i \mid \epsilon_i] \neq \epsilon_i.
\]
Hence, $A_i$ is a biased estimator.

Next, we will show that given two different $\epsilon_i \sim \mathcal{N}(0, \sigma^2)$ and $\epsilon_i' \sim \mathcal{N}(0, \sigma'^2)$, if $\sigma' > \sigma$ then $\text{Bias}(\epsilon_i') >\text{Bias}(\epsilon_i)$ where $\text{Bias}(\epsilon_i) =
\mathbb{E}[A_i \mid \epsilon_i]$.

\textbf{Conditional bias (exact formula)}  

From Step 3, we know that the conditional expectation is 
\[
\mathbb{E}[A_i \mid \epsilon_i] = \left(1 - \frac{1}{N}\right)\epsilon_i \, g(\epsilon_i), 
\qquad 
g(\epsilon_i) = \frac{1}{\sqrt{\mu(\epsilon_i)}} 
+ \frac{3}{8}\frac{\mathrm{Var}(D^2 \mid \epsilon_i)}{\mu(\epsilon_i)^{5/2}}
\]
Hence the \textbf{bias} is
\[
\mathrm{Bias}(\epsilon_i) 
= \mathbb{E}[A_i \mid \epsilon_i] - \epsilon_i 
= \epsilon_i \left[\left(1 - \frac{1}{N}\right) g(\epsilon_i) - 1\right].
\]
Since $\sigma \propto g(\epsilon_i)$ and $g(\epsilon_i) \propto \mathrm{Bias}(\epsilon_i)$, 
we deduce that if $\sigma' > \sigma$, then $\mathrm{Bias}(\epsilon_i') > \mathrm{Bias}(\epsilon_i)$.

\newpage
\section{Hyperparameters Setting}
\label{appendix:verl_hyperparameters}

\begin{table}[!ht]
\centering
\caption{DLER veRL training configuration}
\begin{tabular}{l l}
\toprule
\textbf{Parameter} & \textbf{Value} \\
\midrule
data.train\_batch\_size & 512 \\
actor\_rollout\_ref.actor.ppo\_mini\_batch\_size & 64 \\
actor\_rollout\_ref.actor.ppo\_epochs & 1 \\
data.max\_prompt\_length & 1024 \\
actor\_rollout\_ref.actor.optim.lr & 1.00E-06 \\
actor\_rollout\_ref.rollout.temperature & 1 \\
actor\_rollout\_ref.rollout.n & 16 \\
actor\_rollout\_ref.actor.clip\_ratio\_low & 0.2 \\
actor\_rollout\_ref.actor.clip\_ratio\_high & 0.28 \\
algorithm.filter\_groups.enable & TRUE \\
algorithm.filter\_groups.metric & seq\_reward \\
actor\_rollout\_ref.actor.kl\_loss\_coef & 0.0005 \\
actor\_rollout\_ref.actor.kl\_loss\_type & mse \\
\bottomrule
\end{tabular}
\end{table}

\newpage
\section{Parallel Thinking Latency}
\begin{table}[!htp]\centering
\caption{Average Parallel Inference Latency per Request: DeepSeek-R1-7B vs DLER-R1-7B}\label{tab:parallel_thinking_latency}
\resizebox{ 0.7\textwidth}{!}{ 
\begin{tabular}{lccc}\toprule
& \#Parallel Thinking &Accuracy &Avg Request (Sec) \\\cmidrule{2-4}
\multirow{8}{*}{DeepSeek-R1-1.5B} &1 &29.79 &58.99 \\\cmidrule{2-4}
&4 &53.33 &82.26 \\\cmidrule{2-4}
&8 &63.33 &103.86 \\\cmidrule{2-4}
&12 &67.68 &112.96 \\\cmidrule{2-4}
&16 &71.62 &132.30 \\\cmidrule{2-4}
&32 &78.72 &163.00 \\\cmidrule{2-4}
&64 &80.00 &229.00 \\\cmidrule{1-4}
\multirow{11}{*}{DLER-R1-1.5B} &1 &34.37 &12.35 \\\cmidrule{2-4}
&4 &53.33 &14.83 \\\cmidrule{2-4}
&8 &63.33 &16.64 \\\cmidrule{2-4}
&12 &64.44 &18.91 \\\cmidrule{2-4}
&16 &67.10 &23.70 \\\cmidrule{2-4}
&32 &70.24 &27.64 \\\cmidrule{2-4}
&64 &76.67 &39.99 \\\cmidrule{2-4}
&128 &80.00 &\textbf{52.09}\\\cmidrule{1-4}
\multirow{7}{*}{DeepSeek-R1-7B} &1 &55.40 &93.43 \\\cmidrule{2-4}
&4 &77.75 &144.44 \\\cmidrule{2-4}
&8 &81.28 &174.17 \\\cmidrule{2-4}
&12 &82.47 &200.51 \\\cmidrule{2-4}
&16 &83.33 &221.22 \\\cmidrule{1-4}
\multirow{13}{*}{DLER-R1-7B} &1 &55.60 &23.73 \\\cmidrule{2-4}
&4 &66.67 &26.89 \\\cmidrule{2-4}
&8 &73.20 &29.17 \\\cmidrule{2-4}
&12 &73.88 &32.45 \\\cmidrule{2-4}
&16 &75.17 &34.44 \\\cmidrule{2-4}
&32 &77.69 &43.83 \\\cmidrule{2-4}
&64 &79.56 &55.43 \\\cmidrule{2-4}
&128 &81.67 &57.02 \\\cmidrule{2-4}
&256 &83.33 &\textbf{85.43} \\\midrule
\bottomrule
\end{tabular}}
\end{table}


\end{document}